\documentclass[final]{cvpr}
\usepackage[table, dvipsnames]{xcolor}
\usepackage{times}
\usepackage{epsfig}
\usepackage{graphicx}
\usepackage{amsmath}
\usepackage{amssymb}
\usepackage{subfiles}

\usepackage{makecell} 
\usepackage{wrapfig} 
\usepackage{graphicx}
\usepackage[utf8]{inputenc} %
\usepackage[T1]{fontenc}    %
\usepackage{url}            %
\usepackage{booktabs}       %
\usepackage{amsfonts}       %
\usepackage{nicefrac}       %
\usepackage{microtype}      %
\usepackage{bm}

\usepackage{array}
\usepackage{multirow}

\definecolor{diffcolor}{HTML}{138a08}
\newcommand{\diff}[1]{\color{diffcolor}\footnotesize#1}

\newlength\savewidth

\usepackage{xspace}

\makeatletter
\DeclareRobustCommand\onedot{\futurelet\@let@token\@onedot}
\def\@onedot{\ifx\@let@token.\else.\null\fi\xspace}

\def\eg{\emph{e.g}\onedot} 
\def\ie{\emph{i.e}\onedot} 
 
\def\etc{\emph{etc}\onedot} 
 
\def\etal{\emph{et al}\onedot}
\makeatother

\newcolumntype{S}{>{\centering\arraybackslash}m{0.9cm}}
\newcolumntype{M}{>{\centering\arraybackslash}m{1.2cm}}
\newcolumntype{L}{>{\centering\arraybackslash}m{1.4cm}}
\definecolor{mygray}{gray}{.95}
\definecolor{mylightergray}{gray}{.99}
\definecolor{mygreen}{RGB}{10, 179, 33}
\usepackage{caption} 
\makeatletter
\newcommand{\thickhline}{%
    \noalign {\ifnum 0=`}\fi \hrule height 1pt
    \futurelet \reserved@a \@xhline
}
\newcolumntype{"}{@{\vrule width 1pt}}

\definecolor{citecolor}{HTML}{1a73f2}
\usepackage[pagebackref=true,breaklinks=true,letterpaper=true,colorlinks,
citecolor=citecolor,bookmarks=false]{hyperref}

\def\cvprPaperID{5127} %
\def\confName{ICCV}
\def\confYear{2023}
\graphicspath{{./figs/}{./figs/result/}}

\begin{document}

\title{MotionBERT: A Unified Perspective on Learning Human Motion Representations}

\author{Wentao Zhu\textsuperscript{1} \quad Xiaoxuan Ma\textsuperscript{1} 
\quad Zhaoyang Liu\textsuperscript{2} \quad Libin Liu\textsuperscript{1}
\quad Wayne Wu\textsuperscript{2} \quad Yizhou Wang\textsuperscript{1\footnotemark[2]} \\[1.5ex]
    \textsuperscript{1~}Peking University \qquad
    \textsuperscript{2~}Shanghai AI Laboratory\\[1.1ex]
	{\tt\footnotesize \{wtzhu,maxiaoxuan,libin.liu,yizhou.wang\}@pku.edu.cn}\\	
    {\tt\footnotesize \{zyliumy,wuwenyan0503\}@gmail.com}\\
}

\maketitle
\begin{abstract}

We present a unified perspective on tackling various human-centric video tasks by learning human motion representations from large-scale and heterogeneous data resources. Specifically, we propose a pretraining stage in which a motion encoder is trained to recover the underlying 3D motion from noisy partial 2D observations. The motion representations acquired in this way incorporate geometric, kinematic, and physical knowledge about human motion, which can be easily transferred to multiple downstream tasks. We implement the motion encoder with a Dual-stream Spatio-temporal Transformer (DSTformer) neural network. It could capture long-range spatio-temporal relationships among the skeletal joints comprehensively and adaptively, exemplified by the lowest 3D pose estimation error so far when trained from scratch. Furthermore, our proposed framework achieves state-of-the-art performance on all three downstream tasks by simply finetuning the pretrained motion encoder with a simple regression head (1-2 layers), which demonstrates the versatility of the learned motion representations. Code and models are available at \color{red}{\url{https://motionbert.github.io/}}

\footnotetext[2]{Yizhou Wang is with Center on Frontiers of Computing Studies, School of Computer Science, Peking University and Institute for Artificial Intelligence, Peking University.}

\end{abstract}

\section{Introduction}

\label{sec:1}

Perceiving and understanding human activities have long been a core pursuit of machine intelligence. To this end, researchers define various tasks to estimate \emph{human-centric} semantic labels from videos, \eg skeleton keypoints~\cite{cao2018openpose, alp2018densepose}, action classes~\cite{liu2019ntu, yan2018spatial}, and surface meshes~\cite{loper2015smpl, kocabas2020vibe}. While significant progress has been made in each of these tasks, they tend to be modeled in isolation, rather than as interconnected problems. For example, Spatial Temporal Graph Convolutional Networks (ST-GCN) have been applied to modeling spatio-temporal relationship of human joints in both 3D pose estimation~\cite{cai2019exploiting,wang2020motion} and action recognition~\cite{yan2018spatial, 2sagcn2019cvpr}, but their connections have not been fully explored. Intuitively, these models should all have learned to identify typical human motion patterns, despite being designed for different problems. Nonetheless, current methods fail to mine and utilize such commonalities across the tasks. Ideally, we could develop a unified ~\emph{human-centric} video representation that can be shared across all relevant tasks.

\begin{figure}[t]
  \centering
  \includegraphics[width=\linewidth]{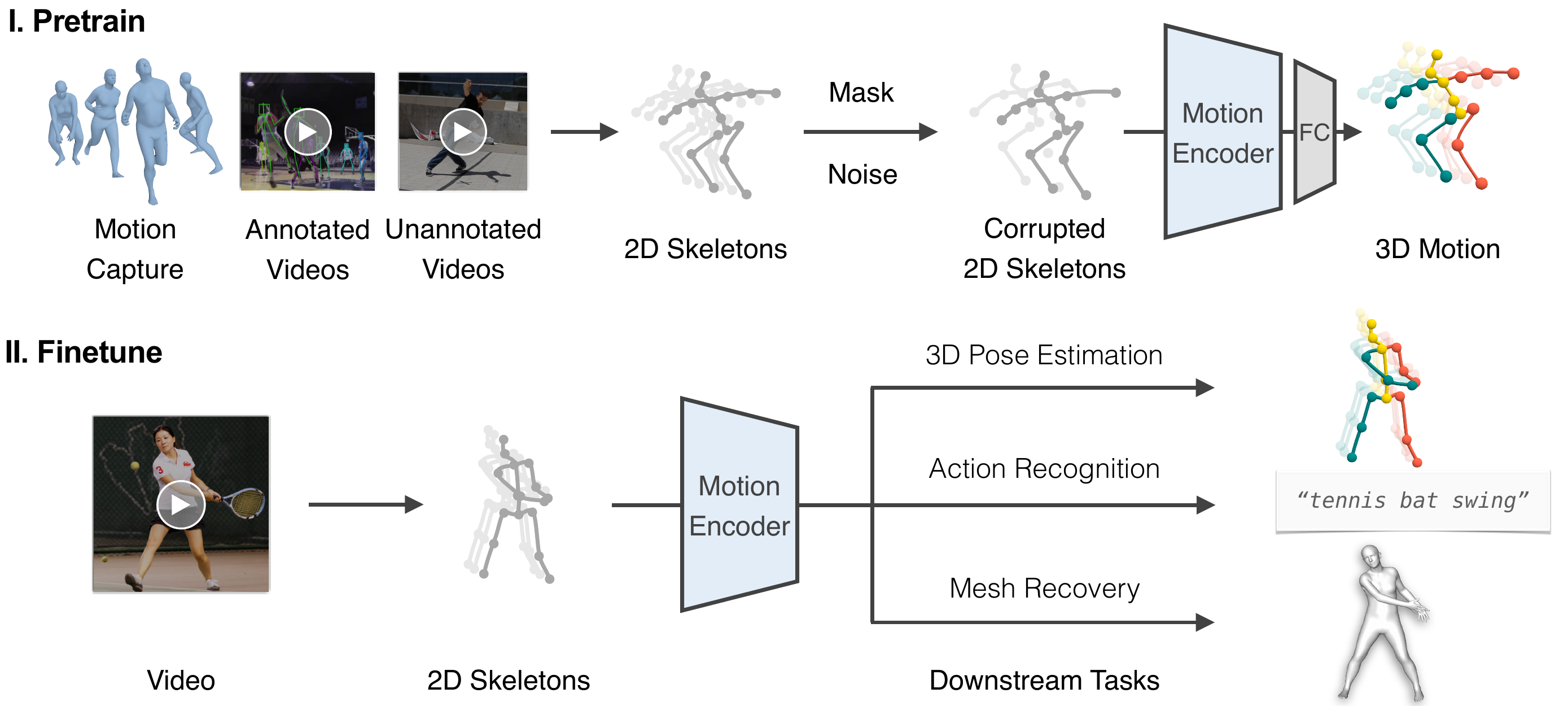}
  \caption{\textbf{Framework overview.} 
  We utilize a motion encoder to learn human motion representations via recovering 3D human motion from corrupted 2D skeleton sequences. To adapt to different downstream tasks, we finetune the pretrained motion representations with a linear layer or a simple MLP. 
  }
  \label{fig:pipeline}
\end{figure}

One significant challenge to developing such a representation is the heterogeneity of available data resources.  Motion capture (Mocap) systems~\cite{h36m_pami, AMASS:2019} provide high-fidelity 3D motion data obtained with markers and sensors, but the appearances of captured videos are usually constrained to simple indoor scenes. Action recognition datasets provide annotations of the action semantics, but they either contain no human pose labels~\cite{shao2020finegym, carreira2017quo} or feature limited motion of daily activities~\cite{liu2017pku, shahroudy2016ntu, liu2019ntu}. In contrast, in-the-wild human videos offer a vast and diverse range of appearance and motion. However, obtaining precise 2D pose annotations requires considerable effort~\cite{PoseTrack}, and acquiring ground-truth (GT) 3D joint locations is almost impossible. Consequently, most existing studies focus on a specific task using a single type of human motion data, and they are not able to enjoy the advantages of other data resources.

In this work, we provide a new perspective on learning human motion representations. The key idea is that we can learn a versatile human motion representation from heterogeneous data resources in a \emph{unified} manner, and utilize the representation to handle different downstream tasks in a \emph{unified} way. We present a two-stage framework, consisting of pretraining and finetuning, as depicted in Figure ~\ref{fig:pipeline}. In the pretraining stage, we extract 2D skeleton sequences from diverse motion data sources and corrupt them with random masks and noises. Subsequently, we train the motion encoder to recover the 3D motion from the corrupted 2D skeletons. This challenging pretext task intrinsically requires the motion encoder to i) infer the underlying 3D human structures from its temporal movements; ii) recover the erroneous and missing observations. In this way, the motion encoder implicitly captures human motion commonsense such as joint linkages, anatomical constraints, and temporal dynamics. In practice, we propose \emph{Dual-stream Spatio-temporal Transformer (DSTformer)} as the motion encoder to capture the long-range relationship among skeleton keypoints. We suppose that the motion representations learned from large-scale and diversified data resources could be shared across different downstream tasks and benefit their performance. Therefore, for each downstream task, we adapt the pretrained motion representations using task-specific training data and supervisory signals with a simple regression head.

In summary, the contributions of this work are three-fold: 1) We provide a new perspective on solving various human-centric video tasks through a shared framework of learning human motion representations. 2) We propose a pretraining method to leverage the large-scale yet heterogeneous human motion resources and learn generalizable human motion representations. Our approach could take advantage of the precision of 3D mocap data and the diversity of in-the-wild RGB videos at the same time. 3) We design a dual-stream Transformer network with cascaded spatio-temporal self-attention blocks that could serve as a general backbone for human motion modeling. The experiments demonstrate that the above designs enable a versatile human motion representation that can be transferred to multiple downstream tasks, outperforming the task-specific state-of-the-art methods.

\section{Related Work}

\paragraph{Learning Human Motion Representations.}  Early works formulate human motion with Hidden Markov Models~\cite{Trabelsi_2013, lehrmann2014efficient} and graphical models~\cite{songmotion2001, kruger2017motionseg}. Kanazawa~\etal~\cite{humanMotionKanazawa19} design a temporal encoder and a hallucinator to learn representations of 3D human dynamics. Zhang~\etal~\cite{Zhang_2019_CVPR} predict future 3D dynamics in a self-supervised manner. Sun~\etal~\cite{sun2021action} further incorporate action labels with an action memory bank. From the action recognition perspective, a variety of pretext tasks are designed to learn motion representations in a self-supervised manner, including future prediction~\cite{su2020predict}, jigsaw puzzle~\cite{lin2020ms2l}, skeleton-contrastive~\cite{Thoker2021skeleton}, speed change~\cite{Su_2021_ICCV}, cross-view consistency~\cite{li2021crossclr}, and contrast-reconstruction~\cite{wang2021contrast}. Similar techniques are also explored in tasks like motion assessment~\cite{Nekoui2021Enhancing, GaitForeMer2022} and motion retargeting~\cite{yang2020transmomo, mocanet2022}. These methods leverage homogeneous motion data, design corresponding pretext tasks, and apply them to a specific downstream task. In this work, we propose a unified pretrain-finetune framework to incorporate heterogeneous data resources and demonstrate its versatility in various downstream tasks.

\paragraph{3D Human Pose Estimation.}
Recovering 3D human poses from monocular RGB videos is a classical problem, and the methods can be categorized into two categories. The first is to estimate 3D poses with CNN directly from images~\cite{DBLP:conf/eccv/SunXWLW18, Moon_2019_ICCV_3DMPPE, ZhouHanICCV19}. However, one limitation of these approaches is that there is a trade-off between 3D pose precision and appearance diversity due to current data collection techniques. The second category is to extract the 2D pose first, then lift the estimated 2D pose to 3D with a separate neural network. The lifting can be achieved via Fully Connected Network~\cite{martinez_2017_3dbaseline, ci2022gfpose}, Temporal Convolutional Network (TCN)~\cite{pavllo20193d, cheng20203d}, GCN~\cite{ci2019optimizing, cai2019exploiting, wang2020motion}, and Transformer~\cite{zheng20213d, li2022mhformer, zhang2022mixste, shan2022p}. Our framework is built upon the second category as we use the proposed DSTformer to accomplish 2D-to-3D lifting.

\paragraph{Skeleton-based Action Recognition.} The pioneering works~\cite{yaoijcv2012, chunyu2013, 8578637} point out the inherent connection between action recognition and human pose estimation. Towards modeling the spatio-temporal relationship among human joints, previous studies mainly employ LSTM~\cite{song2017aaai, zhu2016aaai} and GCN~\cite{yan2018spatial, li2019actional, 2sagcn2019cvpr, liu2020disentangling, cheng2020shiftgcn}. Most recently, PoseConv3D~\cite{duanrevisiting} proposes to apply 3D-CNN on the stacked 2D joint heatmaps and achieves improved results. In addition to the fully-supervised action recognition task, NTU-RGB+D-120~\cite{liu2019ntu} brings attention to the challenging one-shot action recognition problem. To this end, SL-DML~\cite{memmesheimer2021sl} applies deep metric learning to multi-modal signals. Sabater~\etal~\cite{Sabater_2021_CVPR} explores one-shot recognition in therapy scenarios with TCN. We demonstrate that the pretrained motion representations could generalize well to action recognition tasks, and the pretrain-finetune framework is a suitable solution for the one-shot challenges.

\begin{figure*}[t]
  \centering
  \includegraphics[width=0.8\linewidth]{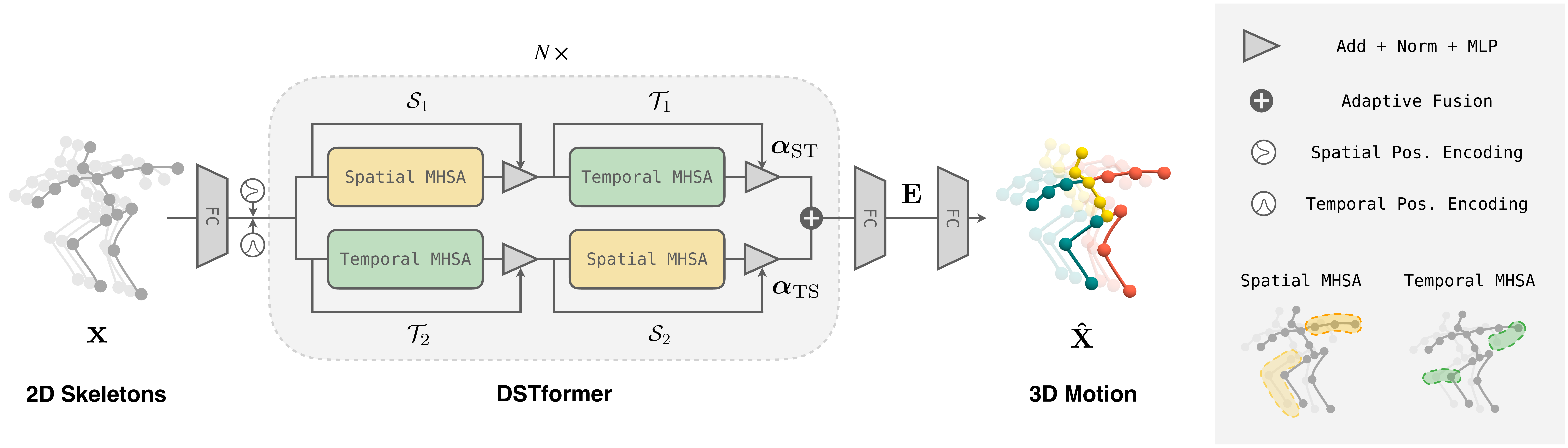}
  \caption{\textbf{Model architecture.} We propose the Dual-stream Spatio-temporal Transformer (DSTformer) as a general backbone for human motion modeling. DSTformer consists of $N$ dual-stream-fusion modules. Each module contains two branches of spatial or temporal MHSA and MLP. The Spatial MHSA models the connection among different joints within a timestep, while the Temporal MHSA models the movement of one joint.}
  \label{fig:arch}
\end{figure*}

\paragraph{Human Mesh Recovery.} Based on the parametric human models such as SMPL~\cite{loper2015smpl}, many research works~\cite{kanazawa2018end, xu2019denserac, moon2020i2l, zhang2021pymaf, Ma_2023_CVPR} focus on regressing the human mesh from a single image. SPIN~\cite{kolotouros2019learning} additionally incorporates fitting the body model to 2D joints in the training loop.
Despite their promising per-frame results, these methods yield jittery and unstable results~\cite{kocabas2020vibe, zeng2022smoothnet} when applied to videos. To improve their temporal coherence, PoseBERT~\cite{baradel2022posebert} and SmoothNet~\cite{zeng2022smoothnet} propose to employ a denoising and smoothing module to the single-frame predictions. Several works~\cite{sun2019human, humanMotionKanazawa19, kocabas2020vibe, choi2021beyond} take video clips as input to exploit the temporal cues. Another common problem is that paired images and GT meshes are mostly captured in constrained scenarios, which limits the generalization ability of the above methods. To that end, Pose2Mesh~\cite{choi2020pose2mesh} proposes to first extract 2D skeletons using an off-the-shelf pose estimator, then lift them to 3D mesh vertices. Our approach is complementary to state-of-the-art human mesh recovery methods and could further improve their temporal coherence with the pretrained motion representations.

\section{Method}
\subsection{Overview}

As discussed in Section~\ref{sec:1},
our approach consists of two stages, namely unified pretraining and task-specific finetuning. In the first stage, we train a motion encoder to accomplish the 2D-to-3D lifting task, where we use the proposed DSTformer as the backbone. In the second stage, we finetune the pretrained motion encoder and a few new layers on the downstream tasks. We use 2D skeleton sequences as input for both pretraining and finetuning because they could be reliably extracted from all kinds of motion sources~\cite{bogo2016keep, AMASS:2019, PoseTrack, newell2016stacked, sun2019deep}, and is more robust to variations~\cite{kwanyee2019weakly3dpose, duanrevisiting}. Existing studies have shown the effectiveness of using 2D skeleton sequences for different downstream tasks~\cite{pavllo20193d, duanrevisiting, choi2020pose2mesh, PoseNet3D2020}. We will first introduce the architecture of DSTformer, and then describe the training scheme in detail.

\subsection{Network Architecture}
Figure \ref{fig:arch} shows the network architecture for 2D-to-3D lifting. Given an input 2D skeleton sequence $\mathbf{x} \in \mathbb{R} ^ {T \times J \times C_{\text{in}}}$, we first project it to a high-dimensional feature $\mathbf{F}^0 \in \mathbb{R} ^ {T \times J \times C_{\text{f}}}$, then add learnable spatial positional encoding $\mathbf{P}_\text{pos}^{\text{S}} \in \mathbb{R}^{1 \times J \times C_{\text{f}}}$ and temporal positional encoding $\mathbf{P}_\text{pos}^{\text{T}} \in \mathbb{R}^{T \times 1 \times C_{\text{f}}}$ to it. We then use the sequence-to-sequence model DSTformer to calculate $\mathbf{F}^{i}\in \mathbb{R}^{T \times J \times C_{\text{f}}}$ ($i = 1,\dots,N$) where $N$ is the network depth. We apply a linear layer with tanh activation \cite{devlin-etal-2019-bert} to $\mathbf{F}^{N}$ to compute the motion representation $\mathbf{E} \in \mathbb{R} ^ {T \times J \times C_{\text{e}}}$.
Finally, we apply a linear transformation to $\mathbf{E}$ to estimate 3D motion $\mathbf{\hat{X}} \in \mathbb{R} ^ {T \times J \times C_{\text{out}}}$. Here, $T$ denotes the sequence length, and $J$ denotes the number of body joints. $C_{\text{in}}$, $C_{\text{f}}$, $C_{\text{e}}$, and $C_{\text{out}}$ denote the channel numbers of input, feature, embedding, and output respectively. We first introduce the basic building blocks of DSTformer, \ie Spatial and Temporal Blocks with Multi-Head Self-Attention (MHSA), and then explain the DSTformer architecture design. 

\label{subsec_arch}

\paragraph{Spatial Block.}
Spatial MHSA (S-MHSA) aims at modeling the relationship among the joints within the same time step. It is defined as 
\begin{equation}
\begin{split}
\small
\text{S-MHSA}(\mathbf{Q}_{\text{S}}, \mathbf{K}_{\text{S}}, \mathbf{V}_{\text{S}})= [\text{head}_{1};...;\text{head}_{h}]\mathbf{W}_\text{S}^{P},  \\
\small
\text{head}_{i}=\text{softmax}( \frac{\mathbf{Q}_\text{S}^{i}(\mathbf{K}_\text{S}^{i})^{\prime}}{\sqrt{d_K}})\mathbf{V}_\text{S}^{i},
\end{split}
\label{eq:multiheadS}
\end{equation}
where $\mathbf{W}_\text{S}^{P}$ is a projection parameter matrix, $h$ is the number of the heads, $i \in 1,\dots,h$, and $^{\prime}$ denotes matrix transpose. We utilize self-attention to get the query $\mathbf{Q}_\text{S}$, key $\mathbf{K}_\text{S}$, and value $\mathbf{V}_\text{S}$ from input per-frame spatial feature $\mathbf{F}_\text{S} \in \mathbb{R} ^ {J \times C_{\text{e}}} $ for each $\text{head}_{i}$,
\begin{align}
~\label{attentionS}
\small
\begin{split}
\mathbf{Q}_\text{S}^{i} = \mathbf{F}_\text{S}\mathbf{W}_\text{S}^{(Q, i)},\ \mathbf{K}_\text{S}^{i} = \mathbf{F}_\text{S}\mathbf{W}_\text{S}^{(K, i)},\
\mathbf{V}_\text{S}^{i} = \mathbf{F}_\text{S}\mathbf{W}_\text{S}^{(V, i)},
\end{split}
\end{align}
where $\mathbf{W}_\text{S}^{(Q, i)}$, $\mathbf{W}_\text{S}^{(K, i)}$, $\mathbf{W}_\text{S}^{(V, i)}$ are projection matrices, and $d_{K}$ is the feature dimension of $\mathbf{K}_\text{S}$. We apply S-MHSA to features of different time steps in parallel. Residual connection and layer normalization (LayerNorm) are used to the S-MHSA result, which is further fed into a multilayer perceptron (MLP), and followed by a residual connection and LayerNorm following~\cite{vaswani2017attention}. We denote the entire spatial block with MHSA, LayerNorm, MLP, and residual connections by $\mathcal{S}$.

\paragraph{Temporal Block.} Temporal MHSA (T-MHSA) aims at modeling the relationship across the time steps for a body joint. Its computation process is similar with S-MHSA except that the MHSA is applied to the per-joint temporal feature $\mathbf{F}_\text{T} \in \mathbb{R} ^ {T \times C_{\text{e}}} $ and parallelized over the spatial dimension. 
\begin{equation}
\begin{split}
\small
\text{T-MHSA}(\mathbf{Q}_\text{T}, \mathbf{K}_\text{T}, \mathbf{V}_\text{T})= [\text{head}_{1};...;\text{head}_{h}]\mathbf{W}_\text{T}^{P}, \\
\small
\text{head}_{i}=\text{softmax}( \frac{\mathbf{Q}_\text{T}^{i}(\mathbf{K}_\text{T}^{i})^{\prime}}{\sqrt{d_K}})\mathbf{V}_\text{T}^{i},
\end{split}
\label{eq:multiheadT}
\end{equation}
where $i \in 1,\dots,h$, $\mathbf{Q}_\text{T}$, $\mathbf{K}_\text{T}$, $\mathbf{V}_\text{T}$ are computed similar with Formula~\ref{attentionS}. We denote the entire temporal block by $\mathcal{T}$.

\paragraph{Dual-stream Spatio-temporal Transformer.} Given spatial and temporal MHSA that captures the intra-frame and inter-frame body joint interactions respectively, we assemble the basic building blocks to fuse the spatial and temporal information in the flow. We design a dual-stream architecture with the following assumptions: 1) Both streams should be capable of modeling the comprehensive spatio-temporal context. 2) Each stream should be specialized in different spatio-temporal aspects. 3) The two streams should be fused together, with the fusion weights dynamically balanced depending on the input spatio-temporal characteristics.

Hence, we stack the spatial and temporal MHSA blocks in different orders, forming two parallel computation branches. The output features of the two branches are fused using adaptive weights predicted by an attention regressor. The dual-stream-fusion module is then repeated for $N$ times:
\begin{equation}
\small
    \mathbf{F}^{i} = \bm{\alpha}_{\text{ST}}^{i} \circ \mathcal{T}^{i}_{1}(\mathcal{S}^{i}_{1} (\mathbf{F}^{i-1})) + \bm{\alpha}_{\text{TS}}^{i} \circ \mathcal{S}^{i}_{2}(\mathcal{T}^{i}_{2}(\mathbf{F}^{i-1})), \quad i \in 1,\dots,N,
\end{equation}
where $\mathbf{F}^{i}$ denotes the feature embedding at depth $i$, $\circ$ denotes element-wise production. Orders of $\mathcal{S}$ and $\mathcal{T}$ blocks are shown in Figure ~\ref{fig:arch}, and different blocks do not share weights. Adaptive fusion weights $\bm{\alpha}_{\text{ST}}, \bm{\alpha}_{\text{TS}} \in \mathbb{R} ^ {N \times T \times J}$ are given by
\begin{equation}
\small
    \bm{\alpha}_{\text{ST}}^{i}, \bm{\alpha}_{\text{TS}}^{i} = \text{softmax}(\mathcal{W}([\mathcal{T}^{i}_{1}(\mathcal{S}^{i}_{1} (\mathbf{F}^{i-1})), \mathcal{S}^{i}_{2}(\mathcal{T}^{i}_{2}(\mathbf{F}^{i-1}))])), 
\end{equation}
where $\mathcal{W}$ is a learnable linear transformation. $[,]$ denotes concatenation.

\subsection{Unified Pretraining}
We address two key challenges when designing the unified pretraining framework: 1) How to learn a powerful motion representation with a universal pretext task. 2) How to utilize large-scale but heterogeneous human motion data in all kinds of formats. 

For the first challenge, we follow the successful practices in language~\cite{devlin-etal-2019-bert, raffel2019exploring, brown2020language} and vision~\cite{he2021mae, bao2022beit} modeling to construct the supervision signals, \ie mask part of the input and use the encoded representations to reconstruct the whole input. Note that such ``cloze'' task naturally exists in human motion analysis, that is to recover the lost depth information from the 2D visual observations, \ie 3D human pose estimation. Inspired by this, we leverage the large-scale 3D mocap data~\cite{AMASS:2019} and design a 2D-to-3D lifting pretext task. We first extract the 2D skeleton sequences $\mathbf{x}$ by projecting the 3D motion orthographically. Then, we corrupt $\mathbf{x}$ by randomly masking and adding noise to produce the corrupted 2D skeleton sequences, which also resemble the 2D detection results as it contains occlusions, detection failures, and errors. Both joint-level and frame-level masks are applied with certain probabilities. We use the aforementioned motion encoder to get motion representation $\mathbf{E}$ and reconstruct 3D motion $\mathbf{\hat{X}}$. We then compute the joint loss $\mathcal{L}_\text{3D}$ between $\mathbf{\hat{X}}$ and GT 3D motion $\mathbf{X}$. We also add the velocity loss $\mathcal{L}_\text{O}$ following previous works~\cite{pavllo20193d, zhang2022mixste}. The 3D reconstruction losses are thus given by 
\begin{equation}
\small
\label{eqn:loss_3d}
    \mathcal{L}_\text{3D} = \sum\limits_{t=1}^{T} \sum\limits_{j=1}^{J} \parallel \mathbf{\hat{X}}_{t,j} - \mathbf{X}_{t,j} \parallel_2, \quad
    \mathcal{L}_\text{O} = \sum\limits_{t=2}^{T} \sum\limits_{j=1}^{J} \parallel \mathbf{\hat{O}}_{t,j} - \mathbf{O}_{t,j} \parallel_2,
\end{equation}
where $\mathbf{\hat{O}}_t=\mathbf{\hat{{X}}}_t - \mathbf{\hat{X}}_{t-1}$, $\mathbf{O}_t=\mathbf{X}_t - \mathbf{X}_{t-1}$.

For the second challenge, we notice that 2D skeletons could serve as a universal medium as they can be extracted from all sorts of motion data sources. We further incorporate in-the-wild RGB videos into the 2D-to-3D lifting framework for unified pretraining. For RGB videos, the 2D skeletons $\mathbf{x}$ could be given by manual annotation~\cite{PoseTrack} or 2D pose estimators~\cite{cao2018openpose, sun2019deep}, and the depth channel of the extracted 2D skeletons is intrinsically ``masked''. Similarly, we add extra masks and noises to degrade $\mathbf{x}$ (if $\mathbf{x}$ already contains detection noise, only masking is applied). As 3D motion GT $\mathbf{X}$ is not available for these data, we apply a weighted 2D re-projection loss which is calculated by
\begin{equation}
\small
    \mathcal{L}_\text{2D} = \sum\limits_{t=1}^{T} \sum\limits_{j=1}^{J} \bm{\delta}_{t,j}  \| \mathbf{\hat{x}}_{t,j} - \mathbf{x}_{t,j} \|_2,
\end{equation}
where $\mathbf{\hat{x}}$ is the 2D orthographical projection of the estimated 3D motion $\mathbf{\hat{X}}$, and $\bm{\delta} \in \mathbb{R} ^ {T \times J}$ is given by visibility annotation or 2D detection confidence. 

The total pretraining loss is computed by
\begin{equation}
\small
    \mathcal{L} = \underbrace{\mathcal{L}_\text{3D} + \lambda_\text{O} \mathcal{L}_\text{O}}_{\text{for 3D data}} + \underbrace{\mathcal{L}_\text{2D}}_{\text{for 2D data}},
\end{equation}
where $\lambda_\text{O}$ is a constant coefficient to balance the losses.

\subsection{Task-specific Finetuning}
The learned feature embedding $\mathbf{E}$ serves as a 3D-aware and temporal-aware human motion representation. For downstream tasks, we adopt the \emph{minimalist} design principle, \ie implementing a shallow downstream network and training without bells and whistles. In practice, we use an extra linear layer or an MLP with one hidden layer. We then finetune the whole network end-to-end. 

\paragraph{3D Pose Estimation.} As we utilize 2D-to-3D lifting as the pretext task, we simply reuse the whole pretrained network. During finetuning, the input 2D skeletons are estimated from videos without extra masks or noises.

\paragraph{Skeleton-based Action Recognition.}
We directly apply a global average pooling over different persons and timesteps. The result is then fed into an MLP with one hidden layer. The network is trained with cross-entropy classification loss. For one-shot learning, we apply a linear layer after the pooled features to extract clip-level action representation. We introduce the detailed setup of one-shot learning in Section~\ref{Sec:experiment-action}.

\paragraph{Human Mesh Recovery.} We use SMPL \cite{loper2015smpl} model to represent the human mesh and regress its parameters. The SMPL model consists of pose parameters $\theta \in \mathbb{R}^{72}$ and shape parameters $\beta \in \mathbb{R}^{10}$, and calculates the 3D mesh as $\mathcal{M}(\theta, \beta) \in \mathbb{R}^{6890 \times 3}$. To regress the pose parameters for each frame, we feed the motion embeddings $\mathbf{E}$ to an MLP with one hidden layer and get $\bm{\hat{\theta}} \in \mathbb{R}^{T \times 72}$. To estimate shape parameters, considering that the human shape over a video sequence is supposed to be consistent, we first perform an average pooling of $\mathbf{E}$ over the temporal dimension and then feed it into another MLP to regress a single $\hat{\beta}$ and then expand it to the entire sequence as $\bm{\hat{\beta}} \in \mathbb{R}^{T \times 10}$. The shape MLP has the same architecture as the pose regression one, and they are initialized with the mean shape and pose, respectively, as in \cite{kocabas2020vibe}. The overall loss is computed as
\begin{equation}
\small
    \mathcal{L} = \lambda^{\text{m}}_\text{3D}\mathcal{L}^{\text{m}}_\text{3D} + \lambda_{\theta} \mathcal{L}_{\theta} + \lambda_{\beta} \mathcal{L}_{\beta} + \lambda_{\text{n}} \mathcal{L}_{\text{norm}} + \lambda^{\text{m}}_\text{O} \mathcal{L}^{\text{m}}_\text{O},
\end{equation}
where each term is calculated as
\begin{equation}
\small
    \begin{aligned}
    & \mathcal{L}^{\text{m}}_\text{3D} = \| \mathbf{\hat{X}}^{\text{m}} - \mathbf{X}^{\text{m}} \|_1, \quad \mathcal{L}_{\theta} = \| \hat{\bm{\theta}} - \bm{\theta} \|_1, \quad
    \mathcal{L}_{\beta} = \| \hat{\bm{\beta}} - \bm{\beta} \|_1,  \\
    & \mathcal{L}_{\text{norm}} = \| \hat{\bm{\theta}}\|_2 + \| \hat{\bm{\beta}}\|_2, \quad
    \mathcal{L}^{\text{m}}_\text{O} =  \| \mathbf{\hat{O}}^{\text{m}} - \mathbf{O}^{\text{m}} \|_2.
    \end{aligned}
\end{equation}
Note that each 3D pose in motion $\mathbf{X}^{\text{m}}$ at frame $t$ is regressed from mesh vertices by $\mathbf{X}^{\text{m}}_t=\mathbf{J} \mathcal{M}(\bm{\theta}_t, \bm{\beta}_t)$, where $\mathbf{J} \in \mathbb{R}^{J \times 6890}$  is a pre-defined matrix \cite{bogo2016keep}. $\mathbf{O}^{\text{m}} = \mathbf{X}^{\text{m}}_{t+1} - \mathbf{X}^{\text{m}}_{t}$, $\hat{\mathbf{O}}^{\text{m}} = \hat{\mathbf{X}}^{\text{m}}_{t+1} - \hat{\mathbf{X}}^{\text{m}}_{t}$. $\lambda^{\text{m}}_\text{3D}$, $\lambda_{\theta}$, $\lambda_{\beta}$, $\lambda_{\text{n}}$ and $\lambda^{\text{m}}_\text{O}$ are constant coefficients to balance the training loss.

\section{Experiments}
\begin{table*}[t]

\center
\small
\setlength{\tabcolsep}{5pt}
\resizebox{0.88\linewidth}{!}{
\begin{tabular}{l c | c c c c c c c c c c c c c c c c}
\thickhline 
Method & $T$ & Dire. & Disc. & Eat & Greet & Phone & Photo & Pose & Purch. & Sit & SitD & Smoke & Wait & WalkD & Walk & WalkT & Avg \\
\hline 

Martinez \etal \cite{martinez_2017_3dbaseline} ICCV'17 & 1& 51.8 & 56.2 & 58.1 & 59.0 & 69.5 & 78.4 & 55.2 & 58.1 & 74.0 & 94.6 & 62.3 & 59.1 & 65.1 & 49.5 & 52.4 & 62.9 \\

Pavlakos \etal \cite{pavlakos2018ordinal} CVPR'18 & 1 & 48.5 & 54.4 & 54.4 & 52.0 & 59.4 & 65.3 & 49.9 & 52.9 & 65.8 & 71.1 & 56.6 & 52.9 & 60.9 & 44.7 & 47.8 & 56.2 \\

LCN \cite{ci2019optimizing} ICCV'19 & 1& 46.8 & 52.3 & 44.7 & 50.4 & 52.9 & 68.9 & 49.6 & 46.4 & 60.2 & 78.9 & 51.2 & 50.0 & 54.8 & 40.4 & 43.3 & 52.7 \\

Xu \etal~\cite{xu2021graph} CVPR'21   & 1&   45.2          & 49.9          & 47.5          & 50.9          & 54.9          & 66.1          & 48.5          & 46.3          & 59.7          & 71.5          & 51.4          & 48.6          & 53.9          & 39.9          & 44.1          & 51.9          \\

\hline

VideoPose3D ~\cite{pavllo20193d} CVPR'19 & 243 & 45.2          & 46.7          & 43.3          & 45.6          & 48.1          & 55.1          & 44.6          & 44.3          & 57.3          & 65.8          & 47.1          & 44.0          & 49.0          & 32.8          & 33.9          & 46.8          \\

Cai \etal~\cite{cai2019exploiting} ICCV'19  & 7  & 44.6          & 47.4          & 45.6          & 48.8          & 50.8          & 59.0          & 47.2          & 43.9          & 57.9          & 61.9          & 49.7          & 46.6          & 51.3          & 37.1          & 39.4          & 48.8          \\

Yeh \etal~\cite{NEURIPS2019_1f88c7c5} NeurIPS'19                     & 243      & 44.8          & 46.1          & 43.3          & 46.4          & 49.0          & 55.2          & 44.6          & 44.0          & 58.3          & 62.7          & 47.1          & 43.9          & 48.6          & 32.7          & 33.3          & 46.7          \\

Liu \etal~\cite{Liu_2020_CVPR} CVPR'20   & 243      & 41.8          & 44.8          & 41.1          & 44.9          & 47.4          & 54.1          & 43.4          & 42.2          & 56.2          & 63.6          & 45.3          & 43.5          & 45.3          & 31.3          & 32.2          & 45.1          \\

 $^\ast$ Cheng \etal~\cite{cheng20203d} AAAI'20         & 128      & \underline{36.2} & \underline{38.1} & 42.7 & 35.9 & \textbf{38.2} & \underline{45.7} & 36.8 & 42.0 & \textbf{45.9} & \textbf{51.3}          & 41.8 & 41.5 & 43.8 & 33.1 & 28.6    & 40.1      \\

$^\ast$ UGCN ~\cite{wang2020motion} ECCV'20                  & 96      & 38.2   & 41.0    & 45.9          & 39.7   & 41.4   & 51.4  & 41.6          & 41.4          & 52.0          & 57.4    & 41.8  & 44.4   & 41.6    & 33.1    & 30.0 &  42.6    \\

$^\dagger$ PoseFormer ~\cite{zheng20213d} ICCV'21          & 81      & 41.5          & 44.8          & 39.8          & 42.5          & 46.5          & 51.6          & 42.1          & 42.0          & 53.3          & 60.7          & 45.5          & 43.3          & 46.1          & 31.8          & 32.2          & 44.3          \\

$^\ast$  Wehrbein \etal~\cite{WehRud2021} ICCV'21         & 200      & 38.5          & 42.5          &  39.9    & 41.7          & 46.5          & 51.6          &  39.9    & 40.8    & 49.5    & 56.8 & 45.3          & 46.4          & 46.8          & 37.8          & 40.4          & 44.3  \\

$^\dagger$ MHFormer ~\cite{li2022mhformer} CVPR'22          & 351      & 39.2          & 43.1          &  40.1   & 40.9          & 44.9          & 51.2          & 40.6   & 41.3    & 53.5    & 60.3 & 43.7          & 41.1          & 43.8          & 29.8          & 30.6         & 43.0  \\

$^\ast$$^\dagger$ MixSTE ~\cite{zhang2022mixste} CVPR'22         &        243    & 36.7 & 39.0 & \underline{36.5} & 39.4 & \underline{40.2} & \textbf{44.9} & 39.8 & 36.9 & 47.9 & 54.8          & \underline{39.6} & 37.8 & 39.3 & 29.7 & 30.6    & 39.8      \\

$^\dagger$  P-STMO ~\cite{shan2022p} ECCV'22        &        243    & 38.4 & 42.1 & 39.8 & 40.2 & 45.2 & 48.9 & 40.4 & 38.3 & 53.8 & 57.3          & 43.9 & 41.6 & 42.2 & 29.3 & 29.3    & 42.1      \\
\hline

\rowcolor{mygray}
$^\dagger$ Ours (scratch)     & 243 & 36.3 & 38.7 & 38.6 & \underline{33.6} & 42.1 & 50.1 & \underline{36.2} & \underline{35.7} & 50.1 & 56.6          & 41.3 & \underline{37.4} & \underline{37.7} & \underline{25.6} & \underline{26.5}    & \underline{39.2}      \\

\rowcolor{mygray}
$^\dagger$ Ours (finetune)     & 243 & \textbf{36.1} & \textbf{37.5} & \textbf{35.8} & \textbf{32.1} & 40.3 & 46.3 & \textbf{36.1} & \textbf{35.3} & \underline{46.9} & \underline{53.9}          & \textbf{39.5} & \textbf{36.3} & \textbf{35.8} & \textbf{25.1} & \textbf{25.3}    & \textbf{37.5}      \\

\thickhline
Method & $T$ & Dire. & Disc. & Eat & Greet & Phone & Photo & Pose & Purch. & Sit & SitD & Smoke & Wait & WalkD & Walk & WalkT & Avg \\
\hline 

Martinez \etal \cite{martinez_2017_3dbaseline} ICCV'17 & 1& 37.7 & 44.4 & 40.3 & 42.1 & 48.2 & 54.9 & 44.4 & 42.1 & 54.6 & 58.0 & 45.1 & 46.4 & 47.6 & 36.4 & 40.4 & 45.5 \\

LCN \cite{ci2019optimizing} ICCV'19 & 1& 36.3 & 38.8 & 29.7 & 37.8 & 34.6 & 42.5 & 39.8 & 32.5 & 36.2 & 39.5 & 34.4 & 38.4 & 38.2 & 31.3 & 34.2 & 36.3 \\

Xu \etal\cite{xu2021graph} CVPR'21   & 1&   35.8          & 38.1          & 31.0          & 35.3          & 35.8          & 43.2          & 37.3          & 31.7          & 38.4          & 45.5          & 35.4          & 36.7          & 36.8          & 27.9          & 30.7          & 35.8          \\

\hline 

UGCN \cite{wang2020motion} ECCV'20             & 96      & 23.0   & 25.7    & 22.8          & 22.6   & 24.1   & 30.6  & 24.9          & 24.5          & 31.1          & 35.0    & 25.6  & 24.3   & 25.1    & 19.8    & 18.4 &  25.6    \\

$^\dagger$ PoseFormer \cite{zheng20213d} ICCV'21      & 81      & 30.0          & 33.6          & 29.9          & 31.0          & 30.2          & 33.3          & 34.8          & 31.4          & 37.8          & 38.6          & 31.7          & 31.5          & 29.0          & 23.3          & 23.1          & 31.3          \\

$^\dagger$ MHFormer \cite{li2022mhformer} CVPR'22       & 351      & 27.7          & 32.1          &  29.1   & 28.9          & 30.0          & 33.9          & 33.0   & 31.2    & 37.0    & 39.3 & 30.0          & 31.0          & 29.4          & 22.2          & 23.0         & 30.5  \\

$^\dagger$ MixSTE \cite{zhang2022mixste}  CVPR'22    &        243    & 21.6 & 22.0 & 20.4 & 21.0 & 20.8 & 24.3 & 24.7 & 21.9 & 26.9 & 24.9          & 21.2 & 21.5 & 20.8 & 14.7 & 15.6    & 21.6      \\

$^\dagger$  P-STMO ~\cite{shan2022p} ECCV'22        &        243    & 28.5 & 30.1 & 28.6 & 27.9  & 29.8 & 33.2 & 31.3 & 27.8 & 36.0 & 37.4 & 29.7 & 29.5 & 28.1 & 21.0 & 21.0 & 29.3      \\
\hline
\rowcolor{mygray}
$^\dagger$ Ours (scratch)     & 243 & \underline{16.7} & \underline{19.9} & \underline{17.1} & \underline{16.5} & \underline{17.4} & \underline{18.8} & \underline{19.3} & \underline{20.5} & \underline{24.0} & \underline{22.1}          & \underline{18.6} & \textbf{16.8} & \underline{16.7} & \underline{10.8} & \underline{11.5}    & \underline{17.8}   \\

\rowcolor{mygray}
 $^\dagger$ Ours (finetune)   & 243 & \textbf{15.9} & \textbf{17.3} & \textbf{16.9} & \textbf{14.6} & \textbf{16.8} & \textbf{18.6} & \textbf{18.6} & \textbf{18.4} & \textbf{22.0} & \textbf{21.8}         & \textbf{17.3} & \underline{16.9} & \textbf{16.1} & \textbf{10.5} & \textbf{11.4}    & \textbf{16.9}      \\

\thickhline

Method & $T$ & Dire. & Disc. & Eat & Greet & Phone & Photo & Pose & Purch. & Sit & SitD & Smoke & Wait & WalkD & Walk & WalkT & Avg \\
\hline 

VideoPose3D ~\cite{pavllo20193d} CVPR'19   & 243      & 3.0          & 3.1          & 2.2          & 3.4          & 2.3          & 2.7          & 2.7          & 3.1          & 2.1          & 2.9          & 2.3          & 2.4          & 3.7          & 3.1          & 2.8          & 2.8          \\

$^\dagger$ PoseFormer \cite{zheng20213d} ICCV'21      & 81      & 3.2          & 3.4          & 2.6          & 3.6          & 2.6          & 3.0          & 2.9          & 3.2          & 2.6          & 3.3          & 2.7          & 2.7          & 3.8          & 3.2          & 2.9          & 3.1          \\

$^\ast$$^\dagger$ MixSTE \cite{zhang2022mixste}  CVPR'22  &        243    & 2.5 & 2.7 & 1.9 & 2.8 & 1.9 & 2.2 & 2.3 & 2.6 & 1.6 & 2.2  & 1.9 & 2.0 & 3.1 & 2.6 & 2.2    & 2.3      \\
\hline
\rowcolor{mygray}
$^\dagger$ Ours (scratch)  & 243 & \underline{1.8} & \underline{2.1} & \underline{1.5} & \underline{2.0} & \underline{1.5} & \underline{1.9} & \underline{1.8} & \underline{2.1} & \underline{1.2} & \underline{1.8}          & \underline{1.5} & \underline{1.4} & \underline{2.6} & \underline{2.0} & \underline{1.7}    & \underline{1.8}   \\

\rowcolor{mygray}
$^\dagger$ Ours (finetune)  & 243 & \textbf{1.7} & \textbf{1.9} & \textbf{1.4} & \textbf{1.9} & \textbf{1.4} & \textbf{1.7} & \textbf{1.7} & \textbf{1.9} & \textbf{1.1} & \textbf{1.6}         & \textbf{1.4} & \textbf{1.3} & \textbf{2.4} & \textbf{1.9} & \textbf{1.6}    & \textbf{1.7}      \\

\thickhline

\end{tabular}}
\vspace{0.2cm}
\caption{\textbf{Quantitative comparison of 3D human pose estimation on Human3.6M.} 
(Top) MPJPE (mm) using detected 2D pose sequences. (Middle) MPJPE (mm) using GT 2D pose sequences. (Bottom) MPJVE (mm) using detected 2D pose sequences. $T$ denotes the clip length used by the method. We select the best results reported by each work. $^\ast$ denotes using HRNet~\cite{sun2019deep} for 2D detection. $^\dagger$ denotes implemented with a spatio-temporal Transformer design. The best and second-best results are highlighted in bold and
underlined formats. 
}
\label{tab:state_of_the_art_h36m}
\end{table*}

\subsection{Implementation}
\label{subsec:implementation}

We implement the proposed motion encoder DSTformer with depth $N=5$, number of heads $h=8$, feature size $C_{\text{f}} = 512$, embedding size $C_{\text{e}} = 512$. For pretraining, we use sequence length $T=243$. The pretrained model could handle different input lengths thanks to the Transformer-based backbone. During finetuning, we set the backbone learning rate to be $0.1\times$ of the new layer learning rate. We introduce the experiment datasets in the following sections respectively. Please refer to the appendix for more experimental details.

\subsection{Pretraining}
\label{Sec:exp-pretrain}
We collect diverse and realistic 3D human motion from two datasets, Human3.6M~\cite{h36m_pami} and AMASS~\cite{AMASS:2019}. Human3.6M~\cite{h36m_pami} is a commonly used indoor dataset for 3D human pose estimation which contains 3.6 million video frames of professional actors performing daily actions. Following previous works~\cite{martinez_2017_3dbaseline, pavllo20193d}, we use subjects 1, 5, 6, 7, 8 for training, and subjects 9, 11 for testing. AMASS~\cite{AMASS:2019} integrates most existing marker-based Mocap datasets~\cite{ACCAD, DanceDB:Aristidou:2019, BMLhandball, BMLrub, cmuWEB, dfaust:CVPR:2017, Eyes_Japan, ghorbani2020movi, chatzitofis2020human4d, HEva_Sigal:IJCV:10b, KIT_Dataset, MoSh_lopermahmoodetal2014, MPI_HDM05, PosePrior_Akhter:CVPR:2015, TCD_hands, DBLP:conf/bmvc/TrumbleGMHC17} and parameterizes them with a common representation. We do not use the images or 2D detection results of the two datasets during pretraining as Mocap datasets usually do not provide raw videos. Instead, we use orthographic projection to get the uncorrupted 2D skeletons. We further incorporate two in-the-wild RGB video datasets PoseTrack~\cite{PoseTrack} (annotated) and InstaVariety~\cite{humanMotionKanazawa19} (unannotated) for higher motion diversity. We align the body keypoint definitions with Human3.6M and calibrate the camera coordinates to pixel coordinates following~\cite{lcn-pami}. We randomly zero out $15\%$ joints, and sample noises from a mixture of Gaussian and uniform distributions~\cite{chang2019poselifter}. We first train on 3D data only for $30$ epochs, then train on both 3D data and 2D data for $60$ epochs, following the curriculum learning practices~\cite{bengio2009icml, 9392296}.

\subsection{3D Pose Estimation}

We evaluate the 3D pose estimation performance on Human3.6M~\cite{h36m_pami} and report the mean per joint position error (MPJPE) in millimeters, which measures the average distance between the predicted joint positions and the GT after aligning the root joint. We also compute the mean per-joint velocity error (MPJVE) to evalute the temporal smoothness following previous works ~\cite{zheng20213d, zhang2022mixste}.
We use the Stacked Hourglass (SH) networks~\cite{newell2016stacked} to extract the 2D skeletons from videos, and finetune the entire network on Human3.6M~\cite{h36m_pami} training set. In addition, we train a separate model of the same architecture, but with random initialization rather than pretrained weights. As shown in Table~\ref{tab:state_of_the_art_h36m} (top), the model trained from scratch outperforms previous methods including other Transformer-based designs with spatio-temporal modeling. It shows the effectiveness of the proposed DSTformer in terms of learning 3D geometric structures and temporal dynamics. To further evaluate the upper bound of the models' capability, we compare the performance when using 2D GT pose sequences as input, which gets rid of the influence of different 2D detectors. As shown in Table~\ref{tab:state_of_the_art_h36m} (middle), our models significantly outperform all the previous approaches. Table~\ref{tab:state_of_the_art_h36m} (bottom) shows that both of our models also surpass previous works in terms of MPJVE, implying better temporal coherence. We attribute the performance advantage of our scratch model to the proposed DSTformer design. We include more comparisons and analysis to demonstrate the advantage of DSTformer with regard to other spatio-temporal architectures in Section~\ref{subsec:ablation} and supplementary materials. Additionally, our method achieves lower errors with the proposed pretraining stage.

\begin{table*}[t]

    \begin{minipage}[t]{0.49\linewidth}\vspace{0pt}%
    \centering
    \begin{center}
    \setlength{\tabcolsep}{3pt}
    \resizebox{0.65\linewidth}{!}{
    \begin{tabular}{c|ccc}
    \thickhline 
    Method & X-Sub & X-View  \\
    \thickhline 
    ST-GCN~\cite{yan2018spatial} AAAI'18 & 81.5 & 88.3  \\

    2s-AGCN~\cite{2sagcn2019cvpr} CVPR'19 & 88.5 & 95.1   \\
    
    MS-G3D~\cite{liu2020disentangling} CVPR'20 & 91.5 & 96.2  \\
    
    Shift-GCN~\cite{cheng2020shiftgcn} CVPR'20 & 90.7 & 96.5  \\
    
    CrosSCLR ~\cite{li2021crossclr} CVPR'21 & 86.2 & 92.5  \\
    
    MCC (finetune)~\cite{Su_2021_ICCV} ICCV'21 & 89.7 & 96.3  \\
    
    SCC (finetune)~\cite{Yang_2021_ICCV} ICCV'21 & 88.0 & 94.9  \\

    UNIK (finetune) ~\cite{yang2021unik} BMVC'21 & 86.8 & 94.4  \\
    CTR-GCN ~\cite{chen2021channel} ICCV'21 & 92.4 & 96.8  \\
    PoseConv3D ~\cite{duanrevisiting} CVPR'22 & \textbf{93.1} & 95.7  \\
    \hline
    \rowcolor{mygray}
    Ours (scratch) & 87.7 & 94.1  \\
    \rowcolor{mygray}
    Ours (finetune) & 93.0 & \textbf{97.2}  \\
    
    \thickhline 
    \end{tabular}
    }
    \end{center}
    \label{tab:nturgbd}
    \end{minipage}
    \begin{minipage}[t]{0.49\linewidth}\vspace{0pt}%
    \centering
    \setlength{\tabcolsep}{18pt}
    \resizebox{0.65\linewidth}{!}{
    \begin{tabular}{c|c}
    \thickhline 
    Method & Accuracy    \\
    \thickhline 
    ST-LSTM + AvgPool~\cite{liu2017skeleton} & 42.9   \\
    ST-LSTM + FC~\cite{liu2017global} & 42.1   \\
    ST-LSTM + Attention~\cite{liu2017global} & 41.0   \\
    APSR~\cite{liu2019ntu} & 45.3 \\
    TCN OneShot~\cite{Sabater_2021_CVPR} & 46.5 \\
    SL-DML~\cite{memmesheimer2021sl} & 50.9  \\
    Skeleton-DML~\cite{Memmesheimer_2022_WACV} & 54.2  \\
    \hline
    \rowcolor{mygray}
    Ours (scratch) & 61.0  \\
    \rowcolor{mygray}
    Ours (finetune) & \textbf{67.4}  \\
    
    \thickhline 
    \end{tabular}
    }
    \label{tab:ntu120_oneshot}
    \end{minipage}
    \caption{\textbf{Quantitative comparison of skeleton-based action recognition accuracy.} (Left) Cross-subject and cross-view recognition accuracy on NTU-RGB+D. All the methods are evaluated using only the ``joint'' modality with 1-clip sampling for the fairness of comparison. (Right) One-shot recognition accuracy on NTU-RGB+D-120. All results are top-1 accuracy ($\%$). 
    }
    \label{tab:action}
\end{table*}

\begin{table*}[ht]
\begin{center}
\setlength{\tabcolsep}{9pt}
\resizebox{0.8\linewidth}{!}{
\small
\begin{tabular}{l|ll|lll|llll}
\thickhline 
\multirow{2}{*}{Method} & \multirow{2}{*}{Input} & \multirow{2}{*}{$T$} & \multicolumn{3}{c}{Human3.6M} & \multicolumn{3}{|c}{3DPW} \\
    \cmidrule[0.05em](lr){4-6} \cmidrule[0.05em](lr){7-9}
    & & & MPVE$\downarrow$ & MPJPE$\downarrow$ & PA-MPJPE$\downarrow$ & MPVE$\downarrow$ & MPJPE$\downarrow$ & PA-MPJPE$\downarrow$ \\
\thickhline 
HMR \cite{kanazawa2018end} CVPR'18 & image & 1 & - & 88.0 & 56.8 & - & 130.0 & 81.3  \\
$^\dagger$ SPIN \cite{kolotouros2019learning} ICCV'19 & image & 1 & 82.3 & 59.4 & 39.3 & 129.1 & 100.9 & 59.1  \\
Pose2Mesh \cite{choi2020pose2mesh} ECCV'20 & 2D pose & 1 & 85.3 & 64.9 & 48.7 & 109.3 & 91.4 & 60.1 \\
I2L-MeshNet \cite{moon2020i2l} ECCV'20 & image & 1 & - & 55.7 & 41.7 & 110.1 & 93.2 & 58.6 \\
$^\dagger$ HybrIK \cite{li2021hybrik} CVPR'21 & image & 1 & 58.1 & 47.4 & 30.1 & 82.4 & 71.3 & 41.9  \\
METRO \cite{lin2021end} CVPR'21 & image & 1 & - & 54.0 & 36.7 & 88.2 & 77.1 & 47.9 \\
Mesh Graphormer\cite{Lin_2021_ICCV} ICCV'21 & image & 1 & - & 51.2 & 34.5 & 87.7 & 74.7 & 45.6 \\
PARE \cite{Kocabas_2021_ICCV} ICCV'21 & image & 1 & - & - & - & 88.6 & 74.5 & 46.5 \\
ROMP \cite{sun2021monocular} ICCV'21 & image & 1 & - & - & - & 108.3 & 91.3 & 54.9 \\
PyMAF \cite{zhang2021pymaf} ICCV'21 & image & 1 & - & 57.7 & 40.5 & 110.1 & 92.8 & 58.9 \\
ProHMR \cite{Kolotouros_2021_ICCV} ICCV'21 & image & 1 & - & - & 41.2 & - & - & 59.8\\
OCHMR \cite{Khirodkar_2022_CVPR} CVPR'22 & image & 1 & - & - & - & 107.1 & 89.7 & 58.3 \\
3DCrowdNet \cite{Choi_2022_CVPR} CVPR'22 & image & 1 & - & - & - & 98.3 & 81.7 & 51.5 \\
CLIFF \cite{li2022cliff} ECCV'22 & image & 1 & - & 47.1 & 32.7 & 81.2 & 69.0 & 43.0 \\
FastMETRO \cite{cho2022FastMETRO} ECCV'22 & image & 1 & - & 52.2 & 33.7 & 84.1 & 73.5 & 44.6 \\
VisDB \cite{yao2022learning} ECCV'22 & image & 1 & - & 51.0 & 34.5 & 85.5 & 73.5 & 44.9 \\
\hline
TemporalContext\cite{arnab2019exploiting} CVPR'19 & video & 32  & - & 77.8 & 54.3  & - & - & 72.2 \\
HMMR \cite{humanMotionKanazawa19} CVPR'19 & video & 20  & - & - & 56.9 & 139.3 & 116.5 & 72.6 \\
DSD-SATN\cite{sun2019human} ICCV'19 & video & 9  & - & 59.1 & 42.4 & - & - & 69.5 \\
VIBE\cite{kocabas2020vibe} CVPR'20 & video & 16  & - & 65.6 & 41.4 & 99.1 & 82.9 & 51.9 \\
TCMR \cite{choi2021beyond} CVPR'21 & video & 16  & - & 62.3 & 41.1 & 102.9 & 86.5 & 52.7 \\
$^\dagger$ MAED \cite{wan2021} ICCV'21 & video & 16  & 84.1 & 60.4 & 38.3 & 93.3 & 79.0 & 45.7 \\
MPS-Net \cite{WeiLin2022mpsnet} CVPR'22 & video & 16  & - & 69.4 & 47.4 & 99.7 & 84.3 & 52.1 \\
$^*$ PoseBERT \cite{baradel2022posebert} TPAMI'22 (+SPIN \cite{kolotouros2019learning}) & video & 16  & - & - & - & - & - & 57.3 \diff{$\downarrow$ 2.3} \\
$^*$ SmoothNet \cite{zeng2022smoothnet} ECCV'22 (+SPIN \cite{kolotouros2019learning}) & video & 32  & - & 67.5 \diff{$\downarrow$ 1.0} & 46.3 \diff{$\downarrow$ 0.2} & - & 86.7 \diff{$\downarrow$ 0.9} & 52.7 \diff{$\downarrow$ 0.6} \\
\hline
\rowcolor{mygray}
Ours (scratch) & 2D motion & 16 & 75.7 & 62.8 & 41.0 & 99.1 & 85.5 & 50.2   \\

\rowcolor{mygray}
Ours (finetune) & 2D motion & 16 & 65.5 & 53.8 & 34.9 & 88.1 & 76.9 & 47.2  \\
\hline
\rowcolor{mygray}
Ours (finetune) + SPIN \cite{kolotouros2019learning} & video & 16  & 63.7 \diff{$\downarrow$ 18.6} & 52.2 \diff{$\downarrow$ 7.2} & 35.7 \diff{$\downarrow$ 3.6} & 92.8 \diff{$\downarrow$36.3} & 79.6 \diff{$\downarrow$ 21.3} & 48.2 \diff{$\downarrow$ 10.9}  \\

\rowcolor{mygray}
Ours (finetune) + MAED \cite{wan2021} & video & 16  & 66.8 \diff{$\downarrow$ 17.3} & 54.8 \diff{$\downarrow$ 5.6} & 36.4 \diff{$\downarrow$ 1.9} & 84.4 \diff{$\downarrow$ 8.9} & 72.3 \diff{$\downarrow$ 6.7} & 42.3 \diff{$\downarrow$ 3.4}  \\

\rowcolor{mygray}
Ours (finetune) + HybrIK \cite{li2021hybrik} & video & 16  & \textbf{52.6} \diff{$\downarrow$ 5.5} & \textbf{43.1} \diff{$\downarrow$ 4.3} & \textbf{27.8} \diff{$\downarrow$ 2.3} & \textbf{79.4} \diff{$\downarrow$ 3.0} & \textbf{68.8} \diff{$\downarrow$ 2.5} & \textbf{40.6} \diff{$\downarrow$ 1.3}  \\

\thickhline 
\end{tabular}
}
\end{center}
\caption{\textbf{Quantitative comparison of human mesh recovery on Human3.6M and 3DPW datasets.} 
$T$ denotes the clip length used by the method.
$^\dagger$ denotes the results obtained with official model weights. The rest are all officially reported results.
The gains in $^*$ correspond to different re-implemented SPIN \cite{kolotouros2019learning} results. 
}
\label{tab:mesh}
\vspace{-0.4cm}
\end{table*}

\subsection{Skeleton-based Action Recognition}
\label{Sec:experiment-action}
We further explore the possibility to learn action semantics with the pretrained human motion representations. We use the human action dataset NTU-RGB+D~\cite{shahroudy2016ntu} which contains 57K videos of 60 action classes, and we follow the data splits Cross-subject (X-Sub) and Cross-view (X-View). The dataset has an extended version, NTU-RGB+D-120~\cite{liu2019ntu}, which contains 114K videos of 120 action classes. We follow the suggested \emph{One-shot} action recognition protocol on NTU-RGB+D-120. For both datasets, we use HRNet~\cite{sun2019deep} to extract 2D skeletons following~\cite{duanrevisiting}. Similarly, we train a scratch model with random initialization for comparison. As Table~\ref{tab:action} (left) shows, our methods are comparable or superior to the state-of-the-art approaches. Notably, the pretraining stage accounts for a large performance gain.

Additionally, we delve into the one-shot setting which holds significant practical importance. Real-world applications often require fine-grained action recognition in specific domains such as education, sports, and healthcare. Unfortunately, the action classes in these scenarios are not typically defined in public datasets. As a result, only limited annotations for these novel action classes are available, making accurate recognition a challenging task. As proposed in ~\cite{liu2019ntu}, we report the results on the evaluation set of $20$ novel classes using only $1$ labeled video for each class. The auxiliary set contains the other $100$ classes, and all samples of these classes can be used. We train the model on the auxiliary set using the supervised contrastive learning technique~\cite{SupContrast}. For a batch of auxiliary data, samples of the same class are pulled together, while samples of different classes are pushed away in the action embedding space. During the evaluation, we calculate the cosine distance between the test examples and the exemplars, and use 1-nearest neighbor to determine the class. Table~\ref{tab:action} (right) illustrates that the proposed models outperform state-of-the-art by a considerable margin. Moreover, it is noteworthy that our pretrained model achieves optimal performance with only 1-2 epochs of fine-tuning. Our results indicate that the pretraining stage is effective in learning a robust motion representation that generalizes well to novel downstream tasks, even with limited data annotations.

\begin{figure*}[t]
  \centering
  \includegraphics[width=\linewidth]{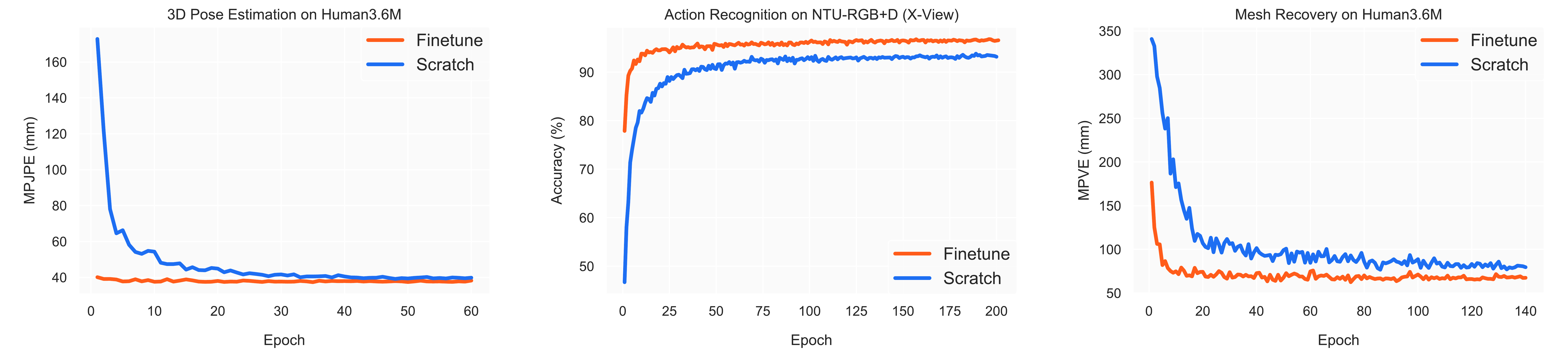}
  \caption{\textbf{Learning curves of finetuning and training from scratch.}}
  \label{fig:training_curve}
\end{figure*}

\subsection{Human Mesh Recovery}
We conduct experiments on Human3.6M \cite{h36m_pami} and 3DPW \cite{vonMarcard2018} datasets and additionally add COCO \cite{lin2014microsoft} dataset during training following \cite{lin2021end, wan2021, kocabas2020vibe}. 
We keep the same training and test split for both datasets as in \cite{martinez_2017_3dbaseline} (Section \ref{Sec:exp-pretrain}) and \cite{kocabas2020vibe, wan2021, lin2021end}, respectively. Following the common practice \cite{kanazawa2018end, kolotouros2019convolutional, kocabas2020vibe, wan2021}, we report MPJPE (mm) and PA-MPJPE (mm) of $14$ joints obtained by $\mathbf{J} \mathcal{M}(\theta, \beta)$. PA-MPJPE calculates MPJPE after aligning with GT in translation, rotation, and scale. We further report the mean per vertex error (MPVE) (mm) of the mesh $\mathcal{M}(\theta, \beta)$, which measures the average distance between the estimated and GT vertices after aligning the root joint. Note that most previous works \cite{kanazawa2018end, kolotouros2019learning, kocabas2020vibe, luo20203d, choi2021beyond, li2021hybrik, lin2021end, wan2021} use more datasets other than COCO~\cite{lin2014microsoft} during training, such as LSP~\cite{johnson2010clustered}, MPI-INF-3DHP~\cite{mehta2017monocular}, \etc, while we do not.
Table \ref{tab:mesh} demonstrates that our finetuned model delivers competitive results on both Human3.6M and 3DPW datasets, surpassing all the state-of-the-art \textit{video-based} methods, including MAED\cite{wan2021}, especially on the MPVE error. Nonetheless, we note that estimating full-body mesh from sparse 2D keypoints alone~\cite{bogo2016keep, choi2020pose2mesh} is an ill-posed problem because it lacks human shape information. In light of this, we propose a hybrid approach that leverages the strengths of both our framework (coherent motion) and RGB-based methods (accurate shape). We introduce a refiner module that can be easily integrated with existing image/video-based methods, similar to ~\cite{baradel2022posebert, zeng2022smoothnet}. Specifically, our refiner module is an MLP that takes the combination of our pretrained motion representations and an initial prediction, regressing a residual in joint rotations. Our approach effectively improves the state-of-the-art methods~\cite{kolotouros2019learning, wan2021, li2021hybrik} and achieves the lowest error to date.

\begin{table}[t]

\center
\small
\resizebox{\linewidth}{!}{
\begin{tabular}{l  | c | c | c | c }
\thickhline 
{Backbone} & {MPJPE $\downarrow$} & MPVE $\downarrow$ & Accuracy $\uparrow$ & Accuracy$\uparrow$  \\
(frozen) & (3D pose)  & (mesh)  & (action x-view)  & (action 1-shot)  \\
\thickhline 

Random & 404.4mm & 114.4mm & 47.6\% & 46.8\%  \\
Pretrained & \textbf{40.3mm} & \textbf{72.1mm} & \textbf{87.3\%} & \textbf{60.7\%}  \\

\thickhline

\end{tabular}}
\vspace{0.1cm}
\caption{\textbf{Comparison of partial finetuning.} 
}
\label{tab:partial}
\vspace{-0.3cm}
\end{table}

\begin{table}[t]
\begin{center}
\resizebox{\linewidth}{!}{
\small
\begin{tabular}{cccc|ccc}
\thickhline 
\multirow{2}{*}{Pretrain} & \multirow{2}{*}{Noise} & \multirow{2}{*}{Mask} & \multirow{2}{*}{2D} & MPJPE$\downarrow$ & MPVE$\downarrow$  &  Accuracy$\uparrow$ \\
& & & & (3D pose)  & (mesh)  & (action x-sub)   \\
\thickhline

 - &     -      &    -        &    -   & 39.2mm & 75.7mm &  87.7\% \\

\checkmark &    -       &      -      &  -   & 38.8mm & 70.6mm &  89.4\% \\
\checkmark & \checkmark &    -       &   -  & 38.1mm & 68.4mm & 90.7\% \\
\checkmark & \checkmark & \checkmark &  -  & \textbf{37.4mm} & 67.8mm & 91.9\% \\
\checkmark & \checkmark & \checkmark &  \checkmark& 37.5mm & \textbf{65.5mm} & \textbf{93.0\%}  \\

\thickhline 
\end{tabular}
}
\end{center}
\caption{\textbf{Comparison of pretraining strategies.}}
\label{tab:ablation_pretrain}
\end{table}

\subsection{Ablation Studies}
\label{subsec:ablation}

\paragraph{Finetune vs. Scratch.} 
We compare the training progress of finetuning the pretrained model and training from scratch. As Figure ~\ref{fig:training_curve} shows, models initialized with pretrained weights demonstrate superior performance and faster convergence on all three tasks. This observation suggests that the pretrained model learns transferable knowledge about human motion, facilitating the learning of multiple downstream tasks.

\paragraph{Partial Finetuning.} 
In addition to end-to-end finetuning, we freeze the motion encoder backbone and only train the regression head for each downstream task. To verify the effectiveness of the pretrained motion representations, we compared the pretrained motion encoder with a randomly initialized motion encoder. We report results of 3D pose and mesh on Human3.6M, action on NTU-RGB+D and NTU-RGB+D-120 (same for the tables below). It can be seen in Table~\ref{tab:partial} that based on the frozen pretrained motion representations, our method still achieves competitive performance on multiple downstream tasks and shows a large improvement compared to the baseline. Pretraining and partial finetuning make it possible for all the downstream tasks to share the same backbone, significantly reducing computation overhead for applications requiring multi-task inference.

\paragraph{Pretraining Strategies.}
We evaluate how different pretraining strategies influence the performance of downstream tasks. Starting from the scratch baseline, we apply the proposed strategies one by one. As shown in Table~\ref{tab:ablation_pretrain}, a vanilla 2D-to-3D pretraining stage brings benefits to all the downstream tasks. Introducing corruptions additionally improves the learned motion embeddings. Unified pretraining with in-the-wild videos (\emph{w.} 2D) enjoys higher motion diversity, which further helps several downstream tasks.

\begin{table}[t]

\center
\small
\resizebox{\linewidth}{!}{
\begin{tabular}{l  | c | c | c | c }
\thickhline 
\multirow{2}{*}{Setting} & {MPJPE $\downarrow$} & MPVE $\downarrow$ & Accuracy $\uparrow$ & Accuracy$\uparrow$  \\
& (3D pose)  & (mesh)  & (action x-view)  & (action 1-shot)  \\
\thickhline 

TCN (scratch) & 50.1mm & 92.6mm & 91.5\% & 52.4\%  \\
TCN (finetune) & \textbf{47.9mm} & \textbf{86.3mm} & \textbf{92.8\%} & \textbf{59.9\%}  \\
\hline 
PoseFormer (scratch) & 44.8mm & 85.9mm & 94.2\% & 57.4\%  \\
PoseFormer (finetune) & \textbf{41.5mm} & \textbf{80.5mm} & \textbf{95.9\%} & \textbf{60.7\%}  \\

\thickhline

\end{tabular}}
\vspace{0.1cm}
\caption{\textbf{Comparison of different backbones.}
}
\label{tab:backbone}
\end{table}

\begin{table}[t]
\begin{center}
\setlength{\tabcolsep}{3pt}
\resizebox{\linewidth}{!}{
\begin{tabular}{c|cccccc}
\thickhline
Arch. & (a)      &  (b)       &  (c)     &  (d) &  (e) & (f) \\
\multirow{2}{*}{Design} & \multirow{2}{*}{S-T}     &  \multirow{2}{*}{T-S}      &  \multirow{2}{*}{S + T}     &  \multirow{2}{*}{ST-MHSA} &  S-T + T-S & S-T + T-S \\
 &   &     &      &   &  (Average) & (Adaptive) \\
\thickhline
MPJPE $\downarrow$ & 40.58$_{\pm0.31}$     & 41.05$_{\pm0.24}$ & 41.76$_{\pm0.22}$       & 41.54$_{\pm0.35}$ & 39.87$_{\pm0.32}$ & \textbf{39.25}$_{\pm0.27}$               \\
 \thickhline
\end{tabular}
}
\end{center}
\caption{\textbf{Comparison of model architecture variants.} All the methods are trained on Human3.6M from scratch over $5$ runs and measured by MPJPE (mm) with mean and standard deviation.}
\label{tab:ablation_arch}
\end{table}

\paragraph{Pretraining with Different Backbones.} 
We further study the universality of the proposed pretraining approach. We replace the motion encoder backbone with two variants: TCN~\cite{pavllo20193d} and PoseFormer~\cite{zheng20213d}. The models are slightly modified to a \textit{seq2seq} version, while all the configurations for pretraining and finetuning are simply followed. Table~\ref{tab:backbone} shows that the proposed approach consistently benefits different backbone models on different tasks.

\paragraph{Model Architecture.} Finally, we study the design choices of DSTformer. From (a) to (f) in Table \ref{tab:ablation_arch}, we compare different structure designs of the basic Transformer module.  (a) and (b) are single-stream versions with different orders. (a) is conceptually similar to PoseFormer~\cite{zheng20213d}, MHFormer~\cite{li2022mhformer}, and MixSTE\cite{zhang2022mixste}. (c) limits each stream to either temporal or spatial modeling before fusion and is similar to MAED~\cite{wan2021}. (d) directly connects S-MHSA and T-MHSA without the MLP in between and is similar to the \textit{MSA-T} variant in MAED~\cite{wan2021}. (e) replaces the adaptive fusion with average pooling on two streams. (f) is the proposed DSTformer design. 
The result statistically confirms our design principles that both streams should be capable and meanwhile complementary, as introduced in Section~\ref{subsec_arch}. In addition, we find out that pairing each self-attention block with an MLP is crucial, as it could project the learned feature interactions and bring nonlinearity. In general, we design the model architecture for the 3D pose estimation task and apply it to all other tasks without additional adjustment.

\section{Conclusion}
\label{Sec:conclusion}
In this work, we provide a unified perspective to tackling various human-centric video tasks. We develop a pretraining approach to learn human motion representations from large-scale and heterogeneous data sources. We also propose DSTformer as a universal human motion encoder. Experimental results on multiple benchmarks demonstrate the versatility of the learned motion representations. Future work could explore fusing the learned motion representations with generic video architectures as a human-centric semantic feature and applying it to more tasks (\eg, action assessment, segmentation).

\noindent\textbf{Acknowledgement}
This work was partially supported by MOST-2022ZD0114900. We would like to thank Hai Ci and Jiefeng Li for their exceptional support.

\clearpage

\bibliographystyle{ieee_fullname}
\bibliography{egbib}

\begin{thebibliography}{100}\itemsep=-1pt

\bibitem{ACCAD}
{Advanced Computing Center for the Arts and Design}.
\newblock {ACCAD MoCap Dataset}.

\bibitem{PosePrior_Akhter:CVPR:2015}
Ijaz Akhter and Michael~J. Black.
\newblock Pose-conditioned joint angle limits for {3D} human pose
  reconstruction.
\newblock In {\em CVPR}, pages 1446--1455, June 2015.

\bibitem{PoseTrack}
Mykhaylo Andriluka, Umar Iqbal, Eldar Insafutdinov, Leonid Pishchulin, Anton
  Milan, Juergen Gall, and Bernt Schiele.
\newblock Posetrack: A benchmark for human pose estimation and tracking.
\newblock In {\em CVPR}, pages 5167--5176, 2018.

\bibitem{andriluka14benchmark}
Mykhaylo Andriluka, Leonid Pishchulin, Peter Gehler, and Bernt Schiele.
\newblock Human pose estimation: New benchmark and state of the art analysis.
\newblock In {\em CVPR}, 2014.

\bibitem{DanceDB:Aristidou:2019}
Andreas Aristidou, Ariel Shamir, and Yiorgos Chrysanthou.
\newblock Digital dance ethnography: Organizing large dance collections.
\newblock {\em JOCCH}, 12(4):1--27, 2019.

\bibitem{arnab2019exploiting}
Anurag Arnab, Carl Doersch, and Andrew Zisserman.
\newblock Exploiting temporal context for 3d human pose estimation in the wild.
\newblock In {\em CVPR}, pages 3395--3404, 2019.

\bibitem{bao2022beit}
Hangbo Bao, Li Dong, Songhao Piao, and Furu Wei.
\newblock {BE}it: {BERT} pre-training of image transformers.
\newblock In {\em ICLR}, 2022.

\bibitem{baradel2022posebert}
Fabien Baradel, Romain Br{\'e}gier, Thibault Groueix, Philippe Weinzaepfel,
  Yannis Kalantidis, and Gr{\'e}gory Rogez.
\newblock Posebert: A generic transformer module for temporal 3d human
  modeling.
\newblock {\em IEEE Transactions on Pattern Analysis and Machine Intelligence},
  2022.

\bibitem{bengio2009icml}
Yoshua Bengio, J\'{e}r\^{o}me Louradour, Ronan Collobert, and Jason Weston.
\newblock Curriculum learning.
\newblock In {\em ICML}, page 41–48, 2009.

\bibitem{bogo2016keep}
Federica Bogo, Angjoo Kanazawa, Christoph Lassner, Peter Gehler, Javier Romero,
  and Michael~J Black.
\newblock Keep it smpl: Automatic estimation of 3d human pose and shape from a
  single image.
\newblock In {\em ECCV}, pages 561--578, 2016.

\bibitem{dfaust:CVPR:2017}
Federica Bogo, Javier Romero, Gerard Pons-Moll, and Michael~J Black.
\newblock Dynamic faust: Registering human bodies in motion.
\newblock In {\em CVPR}, pages 6233--6242, 2017.

\bibitem{brown2020language}
Tom Brown, Benjamin Mann, Nick Ryder, Melanie Subbiah, Jared~D Kaplan, Prafulla
  Dhariwal, Arvind Neelakantan, Pranav Shyam, Girish Sastry, Amanda Askell,
  et~al.
\newblock Language models are few-shot learners.
\newblock {\em NeurIPS}, 33:1877--1901, 2020.

\bibitem{cai2019exploiting}
Yujun Cai, Liuhao Ge, Jun Liu, Jianfei Cai, Tat-Jen Cham, Junsong Yuan, and
  Nadia~Magnenat Thalmann.
\newblock Exploiting spatial-temporal relationships for 3d pose estimation via
  graph convolutional networks.
\newblock In {\em ICCV}, pages 2272--2281, 2019.

\bibitem{cao2018openpose}
Zhe Cao, Tomas Simon, Shih-En Wei, and Yaser Sheikh.
\newblock Realtime multi-person 2d pose estimation using part affinity fields.
\newblock In {\em CVPR}, pages 7291--7299, 2017.

\bibitem{cmuWEB}
{Carnegie Mellon University}.
\newblock {CMU MoCap Dataset}.

\bibitem{carreira2017quo}
Joao Carreira and Andrew Zisserman.
\newblock Quo vadis, action recognition? a new model and the kinetics dataset.
\newblock In {\em CVPR}, pages 6299--6308, 2017.

\bibitem{chang2019poselifter}
Ju~Yong Chang, Gyeongsik Moon, and Kyoung~Mu Lee.
\newblock Poselifter: Absolute 3d human pose lifting network from a single
  noisy 2d human pose.
\newblock {\em arXiv preprint arXiv:1910.12029}, 2019.

\bibitem{chatzitofis2020human4d}
Anargyros Chatzitofis, Leonidas Saroglou, Prodromos Boutis, Petros Drakoulis,
  Nikolaos Zioulis, Shishir Subramanyam, Bart Kevelham, Caecilia Charbonnier,
  Pablo Cesar, Dimitrios Zarpalas, et~al.
\newblock Human4d: A human-centric multimodal dataset for motions and immersive
  media.
\newblock {\em IEEE Access}, 8:176241--176262, 2020.

\bibitem{kwanyee2019weakly3dpose}
Xipeng Chen, Kwan-Yee Lin, Wentao Liu, Chen Qian, and Liang Lin.
\newblock Weakly-supervised discovery of geometry-aware representation for 3d
  human pose estimation.
\newblock In {\em CVPR}, 2019.

\bibitem{chen2021channel}
Yuxin Chen, Ziqi Zhang, Chunfeng Yuan, Bing Li, Ying Deng, and Weiming Hu.
\newblock Channel-wise topology refinement graph convolution for skeleton-based
  action recognition.
\newblock In {\em ICCV}, 2021.

\bibitem{cheng2020shiftgcn}
Ke Cheng, Yifan Zhang, Xiangyu He, Weihan Chen, Jian Cheng, and Hanqing Lu.
\newblock Skeleton-based action recognition with shift graph convolutional
  network.
\newblock In {\em CVPR}, 2020.

\bibitem{cheng20203d}
Yu Cheng, Bo Yang, Bo Wang, and Robby~T Tan.
\newblock 3d human pose estimation using spatio-temporal networks with explicit
  occlusion training.
\newblock In {\em AAAI}, 2020.

\bibitem{cho2022FastMETRO}
Junhyeong Cho, Kim Youwang, and Tae-Hyun Oh.
\newblock Cross-attention of disentangled modalities for 3d human mesh recovery
  with transformers.
\newblock In {\em European Conference on Computer Vision (ECCV)}, 2022.

\bibitem{choi2021beyond}
Hongsuk Choi, Gyeongsik Moon, Ju~Yong Chang, and Kyoung~Mu Lee.
\newblock Beyond static features for temporally consistent 3d human pose and
  shape from a video.
\newblock In {\em CVPR}, pages 1964--1973, 2021.

\bibitem{choi2020pose2mesh}
Hongsuk Choi, Gyeongsik Moon, and Kyoung~Mu Lee.
\newblock Pose2mesh: Graph convolutional network for 3d human pose and mesh
  recovery from a 2d human pose.
\newblock In {\em ECCV}, pages 769--787, 2020.

\bibitem{Choi_2022_CVPR}
Hongsuk Choi, Gyeongsik Moon, JoonKyu Park, and Kyoung~Mu Lee.
\newblock Learning to estimate robust 3d human mesh from in-the-wild crowded
  scenes.
\newblock In {\em CVPR}, pages 1475--1484, June 2022.

\bibitem{lcn-pami}
Hai {Ci}, Xiaoxuan {Ma}, Chunyu {Wang}, and Yizhou {Wang}.
\newblock Locally connected network for monocular 3d human pose estimation.
\newblock {\em IEEE TPAMI}, pages 1--1, 2020.

\bibitem{ci2019optimizing}
Hai Ci, Chunyu Wang, Xiaoxuan Ma, and Yizhou Wang.
\newblock Optimizing network structure for 3d human pose estimation.
\newblock In {\em ICCV}, pages 2262--2271, 2019.

\bibitem{ci2022gfpose}
Hai Ci, Mingdong Wu, Wentao Zhu, Xiaoxuan Ma, Hao Dong, Fangwei Zhong, and
  Yizhou Wang.
\newblock Gfpose: Learning 3d human pose prior with gradient fields.
\newblock {\em CVPR}, 2023.

\bibitem{devlin-etal-2019-bert}
Jacob Devlin, Ming-Wei Chang, Kenton Lee, and Kristina Toutanova.
\newblock {BERT}: Pre-training of deep bidirectional transformers for language
  understanding.
\newblock In {\em NAACL}, 2019.

\bibitem{duan2022PYSKL}
Haodong Duan, Jiaqi Wang, Kai Chen, and Dahua Lin.
\newblock Pyskl: Towards good practices for skeleton action recognition, 2022.

\bibitem{duanrevisiting}
Haodong Duan, Yue Zhao, Kai Chen, Dian Shao, Dahua Lin, and Bo Dai.
\newblock Revisiting skeleton-based action recognition.
\newblock In {\em CVPR}, 2022.

\bibitem{GaitForeMer2022}
Mark Endo, Kathleen~L. Poston, Edith~V. Sullivan, Li Fei-Fei, Kilian~M. Pohl,
  and Ehsan Adeli.
\newblock Gaitforemer: Self-supervised pre-training of transformers via human
  motion forecasting for few-shot gait impairment severity estimation.
\newblock In {\em MICCAI}, pages 130--139, 2022.

\bibitem{ghorbani2020movi}
Saeed Ghorbani, Kimia Mahdaviani, Anne Thaler, Konrad Kording, Douglas~James
  Cook, Gunnar Blohm, and Nikolaus~F. Troje.
\newblock {MoVi}: A large multipurpose motion and video dataset, 2020.

\bibitem{alp2018densepose}
R{\i}za~Alp G{\"u}ler, Natalia Neverova, and Iasonas Kokkinos.
\newblock Densepose: Dense human pose estimation in the wild.
\newblock In {\em CVPR}, pages 7297--7306, 2018.

\bibitem{he2021mae}
Kaiming He, Xinlei Chen, Saining Xie, Yanghao Li, Piotr Doll{\'{a}}r, and
  Ross~B. Girshick.
\newblock Masked autoencoders are scalable vision learners.
\newblock In {\em CVPR}, 2022.

\bibitem{TCD_hands}
Ludovic Hoyet, Kenneth Ryall, Rachel McDonnell, and Carol O’Sullivan.
\newblock Sleight of hand: Perception of finger motion from reduced marker
  sets.
\newblock In {\em Proceedings of the ACM SIGGRAPH Symposium on Interactive 3D
  Graphics and Games}, 2012.

\bibitem{h36m_pami}
Catalin Ionescu, Dragos Papava, Vlad Olaru, and Cristian Sminchisescu.
\newblock Human3.6m: Large scale datasets and predictive methods for 3d human
  sensing in natural environments.
\newblock {\em IEEE TPAMI}, 2014.

\bibitem{johnson2010clustered}
Sam Johnson and Mark Everingham.
\newblock Clustered pose and nonlinear appearance models for human pose
  estimation.
\newblock In {\em BMVC}, volume~2, page~5, 2010.

\bibitem{joo2021exemplar}
Hanbyul Joo, Natalia Neverova, and Andrea Vedaldi.
\newblock Exemplar fine-tuning for 3d human model fitting towards in-the-wild
  3d human pose estimation.
\newblock In {\em 3DV}, pages 42--52. IEEE, 2021.

\bibitem{kanazawa2018end}
Angjoo Kanazawa, Michael~J Black, David~W Jacobs, and Jitendra Malik.
\newblock End-to-end recovery of human shape and pose.
\newblock In {\em CVPR}, pages 7122--7131, 2018.

\bibitem{humanMotionKanazawa19}
Angjoo Kanazawa, Jason~Y Zhang, Panna Felsen, and Jitendra Malik.
\newblock Learning 3d human dynamics from video.
\newblock In {\em CVPR}, pages 5614--5623, 2019.

\bibitem{Khirodkar_2022_CVPR}
Rawal Khirodkar, Shashank Tripathi, and Kris Kitani.
\newblock Occluded human mesh recovery.
\newblock In {\em CVPR}, pages 1715--1725, June 2022.

\bibitem{SupContrast}
Prannay Khosla, Piotr Teterwak, Chen Wang, Aaron Sarna, Yonglong Tian, Phillip
  Isola, Aaron Maschinot, Ce Liu, and Dilip Krishnan.
\newblock Supervised contrastive learning.
\newblock In {\em NeurIPS}, 2020.

\bibitem{kingma2014adam}
Diederik~P Kingma and Jimmy Ba.
\newblock Adam: A method for stochastic optimization.
\newblock {\em arXiv preprint}, arXiv:1412.6980, 2014.

\bibitem{kocabas2020vibe}
Muhammed Kocabas, Nikos Athanasiou, and Michael~J Black.
\newblock Vibe: Video inference for human body pose and shape estimation.
\newblock In {\em CVPR}, pages 5253--5263, 2020.

\bibitem{Kocabas_2021_ICCV}
Muhammed Kocabas, Chun-Hao~P. Huang, Otmar Hilliges, and Michael~J. Black.
\newblock Pare: Part attention regressor for 3d human body estimation.
\newblock In {\em Proceedings of the IEEE/CVF International Conference on
  Computer Vision (ICCV)}, pages 11127--11137, October 2021.

\bibitem{kolotouros2019learning}
Nikos Kolotouros, Georgios Pavlakos, Michael~J Black, and Kostas Daniilidis.
\newblock Learning to reconstruct 3d human pose and shape via model-fitting in
  the loop.
\newblock In {\em ICCV}, pages 2252--2261, 2019.

\bibitem{kolotouros2019convolutional}
Nikos Kolotouros, Georgios Pavlakos, and Kostas Daniilidis.
\newblock Convolutional mesh regression for single-image human shape
  reconstruction.
\newblock In {\em CVPR}, pages 4501--4510, 2019.

\bibitem{Kolotouros_2021_ICCV}
Nikos Kolotouros, Georgios Pavlakos, Dinesh Jayaraman, and Kostas Daniilidis.
\newblock Probabilistic modeling for human mesh recovery.
\newblock In {\em ICCV}, pages 11605--11614, October 2021.

\bibitem{kruger2017motionseg}
Björn Krüger, Anna Vögele, Tobias Willig, Angela Yao, Reinhard Klein, and
  Andreas Weber.
\newblock Efficient unsupervised temporal segmentation of motion data.
\newblock {\em IEEE Transactions on Multimedia}, 2017.

\bibitem{BMLhandball}
Bio~Motion Lab.
\newblock {BMLhandball Motion Capture Database}.

\bibitem{lehrmann2014efficient}
Andreas Lehrmann, Peter~V. Gehler, and Sebastian Nowozin.
\newblock Efficient non-linear markov models for human motion.
\newblock In {\em CVPR}, 2014.

\bibitem{li2021hybrik}
Jiefeng Li, Chao Xu, Zhicun Chen, Siyuan Bian, Lixin Yang, and Cewu Lu.
\newblock Hybrik: A hybrid analytical-neural inverse kinematics solution for 3d
  human pose and shape estimation.
\newblock In {\em CVPR}, pages 3383--3393, 2021.

\bibitem{li2019actional}
Maosen Li, Siheng Chen, Xu Chen, Ya Zhang, Yanfeng Wang, and Qi Tian.
\newblock Actional-structural graph convolutional networks for skeleton-based
  action recognition.
\newblock In {\em CVPR}, 2019.

\bibitem{li2022mhformer}
Wenhao Li, Hong Liu, Hao Tang, Pichao Wang, and Luc Van~Gool.
\newblock Mhformer: Multi-hypothesis transformer for 3d human pose estimation.
\newblock In {\em Proceedings of the IEEE/CVF Conference on Computer Vision and
  Pattern Recognition (CVPR)}, pages 13147--13156, 2022.

\bibitem{li2022cliff}
Zhihao Li, Jianzhuang Liu, Zhensong Zhang, Songcen Xu, and Youliang Yan.
\newblock Cliff: Carrying location information in full frames into human pose
  and shape estimation.
\newblock In {\em ECCV}, 2022.

\bibitem{lin2021end}
Kevin Lin, Lijuan Wang, and Zicheng Liu.
\newblock End-to-end human pose and mesh reconstruction with transformers.
\newblock In {\em CVPR}, pages 1954--1963, 2021.

\bibitem{Lin_2021_ICCV}
Kevin Lin, Lijuan Wang, and Zicheng Liu.
\newblock Mesh graphormer.
\newblock In {\em ICCV}, pages 12939--12948, 2021.

\bibitem{lin2020ms2l}
Lilang Lin, Sijie Song, Wenhan Yang, and Jiaying Liu.
\newblock Ms2l: Multi-task self-supervised learning for skeleton based action
  recognition.
\newblock In {\em ACM International Conference on Multimedia}, pages
  2490--2498, 2020.

\bibitem{lin2014microsoft}
Tsung-Yi Lin, Michael Maire, Serge Belongie, James Hays, Pietro Perona, Deva
  Ramanan, Piotr Doll{\'a}r, and C~Lawrence Zitnick.
\newblock Microsoft coco: Common objects in context.
\newblock In {\em ECCV}, pages 740--755, 2014.

\bibitem{li2021crossclr}
Li Linguo, Wang Minsi, Ni Bingbing, Wang Hang, Yang Jiancheng, and Zhang
  Wenjun.
\newblock 3d human action representation learning via cross-view consistency
  pursuit.
\newblock In {\em CVPR}, 2021.

\bibitem{liu2017pku}
Chunhui Liu, Yueyu Hu, Yanghao Li, Sijie Song, and Jiaying Liu.
\newblock Pku-mmd: A large scale benchmark for continuous multi-modal human
  action understanding.
\newblock {\em arXiv preprint arXiv:1703.07475}, 2017.

\bibitem{liu2019ntu}
Jun Liu, Amir Shahroudy, Mauricio Perez, Gang Wang, Ling-Yu Duan, and Alex~C
  Kot.
\newblock Ntu rgb+ d 120: A large-scale benchmark for 3d human activity
  understanding.
\newblock {\em IEEE TPAMI}, 42(10):2684--2701, 2019.

\bibitem{liu2017skeleton}
Jun Liu, Amir Shahroudy, Dong Xu, Alex~C Kot, and Gang Wang.
\newblock Skeleton-based action recognition using spatio-temporal lstm network
  with trust gates.
\newblock {\em IEEE TPAMI}, 40(12):3007--3021, 2017.

\bibitem{liu2017global}
Jun Liu, Gang Wang, Ping Hu, Ling-Yu Duan, and Alex~C Kot.
\newblock Global context-aware attention lstm networks for 3d action
  recognition.
\newblock In {\em CVPR}, pages 1647--1656, 2017.

\bibitem{Liu_2020_CVPR}
Ruixu Liu, Ju Shen, He Wang, Chen Chen, Sen-ching Cheung, and Vijayan Asari.
\newblock Attention mechanism exploits temporal contexts: Real-time 3d human
  pose reconstruction.
\newblock In {\em CVPR}, 2020.

\bibitem{liu2020disentangling}
Ziyu Liu, Hongwen Zhang, Zhenghao Chen, Zhiyong Wang, and Wanli Ouyang.
\newblock Disentangling and unifying graph convolutions for skeleton-based
  action recognition.
\newblock In {\em CVPR}, pages 143--152, 2020.

\bibitem{MoSh_lopermahmoodetal2014}
Matthew Loper, Naureen Mahmood, and Michael~J. Black.
\newblock {MoSh}: {Motion} and {Shape Capture} from {Sparse Markers}.
\newblock {\em ACM Transactions on Graphics}, 33(6), Nov. 2014.

\bibitem{loper2014mosh}
Matthew Loper, Naureen Mahmood, and Michael~J Black.
\newblock Mosh: Motion and shape capture from sparse markers.
\newblock {\em TOG}, 33(6):1--13, 2014.

\bibitem{loper2015smpl}
Matthew Loper, Naureen Mahmood, Javier Romero, Gerard Pons-Moll, and Michael~J
  Black.
\newblock Smpl: A skinned multi-person linear model.
\newblock {\em ACM Transactions on Graphics}, 34(6):1--16, 2015.

\bibitem{Eyes_Japan}
Eyes JAPAN~Co. Ltd.
\newblock {Eyes Japan MoCap Dataset}.

\bibitem{luo20203d}
Zhengyi Luo, S~Alireza Golestaneh, and Kris~M Kitani.
\newblock 3d human motion estimation via motion compression and refinement.
\newblock In {\em ACCV}, 2020.

\bibitem{8578637}
Diogo~C. Luvizon, David Picard, and Hedi Tabia.
\newblock 2d/3d pose estimation and action recognition using multitask deep
  learning.
\newblock In {\em CVPR}, pages 5137--5146, 2018.

\bibitem{Ma_2023_CVPR}
Xiaoxuan Ma, Jiajun Su, Chunyu Wang, Wentao Zhu, and Yizhou Wang.
\newblock 3d human mesh estimation from virtual markers.
\newblock In {\em Proceedings of the IEEE/CVF Conference on Computer Vision and
  Pattern Recognition (CVPR)}, pages 534--543, June 2023.

\bibitem{AMASS:2019}
Naureen Mahmood, Nima Ghorbani, Nikolaus~F. Troje, Gerard Pons-Moll, and
  Michael~J. Black.
\newblock {AMASS}: Archive of motion capture as surface shapes.
\newblock In {\em ICCV}, pages 5441--5450, Oct. 2019.

\bibitem{KIT_Dataset}
Christian Mandery, {\"O}mer Terlemez, Martin Do, Nikolaus Vahrenkamp, and Tamim
  Asfour.
\newblock The kit whole-body human motion database.
\newblock In {\em ICAR}, pages 329--336. IEEE, 2015.

\bibitem{martinez_2017_3dbaseline}
Julieta Martinez, Rayat Hossain, Javier Romero, and James~J Little.
\newblock A simple yet effective baseline for 3d human pose estimation.
\newblock In {\em ICCV}, pages 2640--2649, 2017.

\bibitem{mehta2017monocular}
Dushyant Mehta, Helge Rhodin, Dan Casas, Pascal Fua, Oleksandr Sotnychenko,
  Weipeng Xu, and Christian Theobalt.
\newblock Monocular 3d human pose estimation in the wild using improved cnn
  supervision.
\newblock In {\em 3DV}, pages 506--516. IEEE, 2017.

\bibitem{Memmesheimer_2022_WACV}
Raphael Memmesheimer, Simon H\"aring, Nick Theisen, and Dietrich Paulus.
\newblock Skeleton-dml: Deep metric learning for skeleton-based one-shot action
  recognition.
\newblock In {\em WACV}, pages 3702--3710, 2022.

\bibitem{memmesheimer2021sl}
Raphael Memmesheimer, Nick Theisen, and Dietrich Paulus.
\newblock Sl-dml: Signal level deep metric learning for multimodal one-shot
  action recognition.
\newblock In {\em ICPR}, 2021.

\bibitem{Moon_2019_ICCV_3DMPPE}
Gyeongsik Moon, Juyong Chang, and Kyoung~Mu Lee.
\newblock Camera distance-aware top-down approach for 3d multi-person pose
  estimation from a single rgb image.
\newblock In {\em ICCV}, 2019.

\bibitem{moon2020i2l}
Gyeongsik Moon and Kyoung~Mu Lee.
\newblock I2l-meshnet: Image-to-lixel prediction network for accurate 3d human
  pose and mesh estimation from a single rgb image.
\newblock In {\em ECCV}, pages 752--768, 2020.

\bibitem{MPI_HDM05}
Meinard M{\"u}ller, Tido R{\"o}der, Michael Clausen, Bernhard Eberhardt,
  Bj{\"o}rn Kr{\"u}ger, and Andreas Weber.
\newblock Documentation mocap database {HDM05}.
\newblock Technical Report CG-2007-2, Universit\"{a}t Bonn, June 2007.

\bibitem{Nekoui2021Enhancing}
Mahdiar Nekoui and Li Cheng.
\newblock Enhancing human motion assessment by self-supervised representation
  learning.
\newblock In {\em BMVC}, 2021.

\bibitem{newell2016stacked}
Alejandro Newell, Kaiyu Yang, and Jia Deng.
\newblock Stacked hourglass networks for human pose estimation.
\newblock In {\em ECCV}, 2016.

\bibitem{NEURIPS2019_9015}
Adam Paszke, Sam Gross, Francisco Massa, Adam Lerer, James Bradbury, Gregory
  Chanan, Trevor Killeen, Zeming Lin, Natalia Gimelshein, Luca Antiga, Alban
  Desmaison, Andreas Kopf, Edward Yang, Zachary DeVito, Martin Raison, Alykhan
  Tejani, Sasank Chilamkurthy, Benoit Steiner, Lu Fang, Junjie Bai, and Soumith
  Chintala.
\newblock Pytorch: An imperative style, high-performance deep learning library.
\newblock In {\em NeurIPS}. 2019.

\bibitem{pavlakos2018ordinal}
Georgios Pavlakos, Xiaowei Zhou, and Kostas Daniilidis.
\newblock Ordinal depth supervision for 3d human pose estimation.
\newblock In {\em CVPR}, pages 7307--7316, 2018.

\bibitem{pavllo20193d}
Dario Pavllo, Christoph Feichtenhofer, David Grangier, and Michael Auli.
\newblock 3d human pose estimation in video with temporal convolutions and
  semi-supervised training.
\newblock In {\em CVPR}, pages 7753--7762, 2019.

\bibitem{raffel2019exploring}
Colin Raffel, Noam Shazeer, Adam Roberts, Katherine Lee, Sharan Narang, Michael
  Matena, Yanqi Zhou, Wei Li, and Peter~J Liu.
\newblock Exploring the limits of transfer learning with a unified text-to-text
  transformer.
\newblock {\em Journal of Machine Learning Research}, 2020.

\bibitem{MANO:SIGGRAPHASIA:2017}
Javier Romero, Dimitrios Tzionas, and Michael~J. Black.
\newblock Embodied hands: Modeling and capturing hands and bodies together.
\newblock {\em ACM Transactions on Graphics, (Proc. SIGGRAPH Asia)}, 2017.

\bibitem{Sabater_2021_CVPR}
Alberto Sabater, Laura Santos, Jose Santos-Victor, Alexandre Bernardino, Luis
  Montesano, and Ana~C. Murillo.
\newblock One-shot action recognition in challenging therapy scenarios.
\newblock In {\em CVPR Workshop}, 2021.

\bibitem{shahroudy2016ntu}
Amir Shahroudy, Jun Liu, Tian-Tsong Ng, and Gang Wang.
\newblock Ntu rgb+ d: A large scale dataset for 3d human activity analysis.
\newblock In {\em CVPR}, pages 1010--1019, 2016.

\bibitem{shan2022p}
Wenkang Shan, Zhenhua Liu, Xinfeng Zhang, Shanshe Wang, Siwei Ma, and Wen Gao.
\newblock P-stmo: Pre-trained spatial temporal many-to-one model for 3d human
  pose estimation.
\newblock In {\em Computer Vision--ECCV 2022: 17th European Conference, Tel
  Aviv, Israel, October 23--27, 2022, Proceedings, Part V}, pages 461--478.
  Springer, 2022.

\bibitem{shao2020finegym}
Dian Shao, Yue Zhao, Bo Dai, and Dahua Lin.
\newblock Finegym: A hierarchical video dataset for fine-grained action
  understanding.
\newblock In {\em CVPR}, pages 2616--2625, 2020.

\bibitem{2sagcn2019cvpr}
Lei Shi, Yifan Zhang, Jian Cheng, and Hanqing Lu.
\newblock Two-stream adaptive graph convolutional networks for skeleton-based
  action recognition.
\newblock In {\em CVPR}, 2019.

\bibitem{HEva_Sigal:IJCV:10b}
Leonid Sigal, Alexandru~O Balan, and Michael~J Black.
\newblock Humaneva: Synchronized video and motion capture dataset and baseline
  algorithm for evaluation of articulated human motion.
\newblock {\em IJCV}, 87(1):4--27, 2010.

\bibitem{song2017aaai}
Sijie Song, Cuiling Lan, Junliang Xing, Wenjun Zeng, and Jiaying Liu.
\newblock An end-to-end spatio-temporal attention model for human action
  recognition from skeleton data.
\newblock In {\em AAAI}, page 4263–4270, 2017.

\bibitem{songmotion2001}
Yang Song, Luis Goncalves, and Pietro Perona.
\newblock Unsupervised learning of human motion models.
\newblock In T. Dietterich, S. Becker, and Z. Ghahramani, editors, {\em
  NeurIPS}, volume~14. MIT Press, 2001.

\bibitem{su2020predict}
Kun Su, Xiulong Liu, and Eli Shlizerman.
\newblock Predict \& cluster: Unsupervised skeleton based action recognition.
\newblock In {\em CVPR}, pages 9631--9640, 2020.

\bibitem{Su_2021_ICCV}
Yukun Su, Guosheng Lin, and Qingyao Wu.
\newblock Self-supervised 3d skeleton action representation learning with
  motion consistency and continuity.
\newblock In {\em ICCV}, 2021.

\bibitem{sun2021action}
Jiangxin Sun, Zihang Lin, Xintong Han, Jian-Fang Hu, Jia Xu, and Wei-Shi Zheng.
\newblock Action-guided 3d human motion prediction.
\newblock In {\em NeurIPS}, volume~34, pages 30169--30180. Curran Associates,
  Inc., 2021.

\bibitem{sun2019deep}
Ke Sun, Bin Xiao, Dong Liu, and Jingdong Wang.
\newblock Deep high-resolution representation learning for human pose
  estimation.
\newblock In {\em CVPR}, 2019.

\bibitem{DBLP:conf/eccv/SunXWLW18}
Xiao Sun, Bin Xiao, Fangyin Wei, Shuang Liang, and Yichen Wei.
\newblock Integral human pose regression.
\newblock In {\em ECCV}, 2018.

\bibitem{sun2021monocular}
Yu Sun, Qian Bao, Wu Liu, Yili Fu, Michael~J Black, and Tao Mei.
\newblock Monocular, one-stage, regression of multiple 3d people.
\newblock In {\em ICCV}, pages 11179--11188, 2021.

\bibitem{sun2019human}
Yu Sun, Yun Ye, Wu Liu, Wenpeng Gao, Yili Fu, and Tao Mei.
\newblock Human mesh recovery from monocular images via a skeleton-disentangled
  representation.
\newblock In {\em ICCV}, pages 5349--5358, 2019.

\bibitem{Thoker2021skeleton}
Fida~Mohammad Thoker, Hazel Doughty, and Cees~GM Snoek.
\newblock Skeleton-contrastive 3d action representation learning.
\newblock In {\em ACM International Conference on Multimedia}, pages
  1655--1663, 2021.

\bibitem{Trabelsi_2013}
Dorra Trabelsi, Samer Mohammed, Faicel Chamroukhi, Latifa Oukhellou, and Yacine
  Amirat.
\newblock An unsupervised approach for automatic activity recognition based on
  hidden markov model regression.
\newblock {\em {IEEE} Transactions on Automation Science and Engineering},
  2013.

\bibitem{PoseNet3D2020}
Shashank Tripathi, Siddhant Ranade, Ambrish Tyagi, and Amit~K. Agrawal.
\newblock Posenet3d: Learning temporally consistent 3d human pose via knowledge
  distillation.
\newblock In {\em 8th International Conference on 3D Vision}, 2020.

\bibitem{BMLrub}
Nikolaus~F. Troje.
\newblock Decomposing biological motion: {A} framework for analysis and
  synthesis of human gait patterns.
\newblock {\em Journal of Vision}, 2(5):2--2, Sept. 2002.

\bibitem{DBLP:conf/bmvc/TrumbleGMHC17}
Matthew Trumble, Andrew Gilbert, Charles Malleson, Adrian Hilton, and John~P.
  Collomosse.
\newblock Total capture: 3d human pose estimation fusing video and inertial
  sensors.
\newblock In {\em BMVC}, 2017.

\bibitem{vaswani2017attention}
Ashish Vaswani, Noam Shazeer, Niki Parmar, Jakob Uszkoreit, Llion Jones,
  Aidan~N Gomez, \L~ukasz Kaiser, and Illia Polosukhin.
\newblock Attention is all you need.
\newblock In {\em NeurIPS}, 2017.

\bibitem{vonMarcard2018}
Timo von Marcard, Roberto Henschel, Michael~J Black, Bodo Rosenhahn, and Gerard
  Pons-Moll.
\newblock Recovering accurate 3d human pose in the wild using imus and a moving
  camera.
\newblock In {\em ECCV}, pages 601--617, 2018.

\bibitem{wan2021}
Ziniu Wan, Zhengjia Li, Maoqing Tian, Jianbo Liu, Shuai Yi, and Hongsheng Li.
\newblock Encoder-decoder with multi-level attention for 3d human shape and
  pose estimation.
\newblock In {\em ICCV}, pages 13033--13042, 2021.

\bibitem{chunyu2013}
Chunyu Wang, Yizhou Wang, and Alan~L. Yuille.
\newblock An approach to pose-based action recognition.
\newblock In {\em CVPR}, pages 915--922, 2013.

\bibitem{wang2020motion}
Jingbo Wang, Sijie Yan, Yuanjun Xiong, and Dahua Lin.
\newblock Motion guided 3d pose estimation from videos.
\newblock In {\em European Conference on Computer Vision}, pages 764--780.
  Springer, 2020.

\bibitem{wang2021contrast}
Peng Wang, Jun Wen, Chenyang Si, Yuntao Qian, and Liang Wang.
\newblock Contrast-reconstruction representation learning for self-supervised
  skeleton-based action recognition.
\newblock {\em arXiv preprint arXiv:2111.11051}, 2021.

\bibitem{9392296}
Xin Wang, Yudong Chen, and Wenwu Zhu.
\newblock A survey on curriculum learning.
\newblock {\em IEEE TPAMI}, 2022.

\bibitem{WehRud2021}
Tom Wehrbein, Marco Rudolph, Bodo Rosenhahn, and Bastian Wandt.
\newblock Probabilistic monocular 3d human pose estimation with normalizing
  flows.
\newblock In {\em ICCV}, 2021.

\bibitem{WeiLin2022mpsnet}
Wen-Li Wei, Jen-Chun Lin, Tyng-Luh Liu, and Hong-Yuan~Mark Liao.
\newblock Capturing humans in motion: Temporal-attentive 3d human pose and
  shape estimation from monocular video.
\newblock In {\em The IEEE Conference on Computer Vision and Pattern
  Recognition (CVPR)}, June 2022.

\bibitem{xu2021graph}
Tianhan Xu and Wataru Takano.
\newblock Graph stacked hourglass networks for 3d human pose estimation.
\newblock In {\em CVPR}, 2021.

\bibitem{xu2019denserac}
Yuanlu Xu, Song-Chun Zhu, and Tony Tung.
\newblock Denserac: Joint 3d pose and shape estimation by dense
  render-and-compare.
\newblock In {\em ICCV}, pages 7760--7770, 2019.

\bibitem{yan2018spatial}
Sijie Yan, Yuanjun Xiong, and Dahua Lin.
\newblock Spatial temporal graph convolutional networks for skeleton-based
  action recognition.
\newblock In {\em AAAI}, 2018.

\bibitem{yang2021unik}
Di Yang, Yaohui Wang, Antitza Dantcheva, Lorenzo Garattoni, Gianpiero
  Francesca, and Francois Bremond.
\newblock Unik: A unified framework for real-world skeleton-based action
  recognition.
\newblock In {\em BMVC}, 2021.

\bibitem{Yang_2021_ICCV}
Siyuan Yang, Jun Liu, Shijian Lu, Meng~Hwa Er, and Alex~C. Kot.
\newblock Skeleton cloud colorization for unsupervised 3d action representation
  learning.
\newblock In {\em ICCV}, 2021.

\bibitem{yang2020transmomo}
Zhuoqian Yang, Wentao Zhu, Wayne Wu, Chen Qian, Qiang Zhou, Bolei Zhou, and
  Chen~Change Loy.
\newblock Transmomo: Invariance-driven unsupervised video motion retargeting.
\newblock In {\em CVPR}, 2020.

\bibitem{yaoijcv2012}
Angela Yao, Juergen Gall, and Luc Gool.
\newblock Coupled action recognition and pose estimation from multiple views.
\newblock {\em Int. J. Comput. Vision}, page 16–37, 2012.

\bibitem{yao2022learning}
Chun-Han Yao, Jimei Yang, Duygu Ceylan, Yi Zhou, Yang Zhou, and Ming-Hsuan
  Yang.
\newblock Learning visibility for robust dense human body estimation.
\newblock In {\em European conference on computer vision (ECCV)}, 2022.

\bibitem{NEURIPS2019_1f88c7c5}
Raymond Yeh, Yuan-Ting Hu, and Alexander Schwing.
\newblock Chirality nets for human pose regression.
\newblock In {\em NeurIPS}, 2019.

\bibitem{zeng2022smoothnet}
Ailing Zeng, Lei Yang, Xuan Ju, Jiefeng Li, Jianyi Wang, and Qiang Xu.
\newblock Smoothnet: A plug-and-play network for refining human poses in
  videos.
\newblock In {\em European Conference on Computer Vision}. Springer, 2022.

\bibitem{Zhang_2020_CVPR}
Feng Zhang, Xiatian Zhu, Hanbin Dai, Mao Ye, and Ce Zhu.
\newblock Distribution-aware coordinate representation for human pose
  estimation.
\newblock In {\em CVPR}, June 2020.

\bibitem{Zhang_2019_CVPR}
Feng Zhang, Xiatian Zhu, and Mao Ye.
\newblock Fast human pose estimation.
\newblock In {\em CVPR}, 2019.

\bibitem{zhang2021pymaf}
Hongwen Zhang, Yating Tian, Xinchi Zhou, Wanli Ouyang, Yebin Liu, Limin Wang,
  and Zhenan Sun.
\newblock Pymaf: 3d human pose and shape regression with pyramidal mesh
  alignment feedback loop.
\newblock In {\em ICCV}, pages 11446--11456, 2021.

\bibitem{zhang2022mixste}
Jinlu Zhang, Zhigang Tu, Jianyu Yang, Yujin Chen, and Junsong Yuan.
\newblock Mixste: Seq2seq mixed spatio-temporal encoder for 3d human pose
  estimation in video.
\newblock In {\em CVPR}, 2022.

\bibitem{zheng20213d}
Ce Zheng, Sijie Zhu, Matias Mendieta, Taojiannan Yang, Chen Chen, and Zhengming
  Ding.
\newblock 3d human pose estimation with spatial and temporal transformers.
\newblock {\em ICCV}, 2021.

\bibitem{ZhouHanICCV19}
Kun Zhou, Xiaoguang Han, Nianjuan Jiang, Kui Jia, and Jiangbo Lu.
\newblock Hemlets pose: Learning part-centric heatmap triplets for accurate 3d
  human pose estimation.
\newblock In {\em ICCV}, 2019.

\bibitem{zhou2019continuity}
Yi Zhou, Connelly Barnes, Jingwan Lu, Jimei Yang, and Hao Li.
\newblock On the continuity of rotation representations in neural networks.
\newblock In {\em CVPR}, 2019.

\bibitem{zhu2016aaai}
Wentao Zhu, Cuiling Lan, Junliang Xing, Wenjun Zeng, Yanghao Li, Li Shen, and
  Xiaohui Xie.
\newblock Co-occurrence feature learning for skeleton based action recognition
  using regularized deep lstm networks.
\newblock In {\em AAAI}, 2016.

\bibitem{mocanet2022}
Wentao Zhu, Zhuoqian Yang, Ziang Di, Wayne Wu, Yizhou Wang, and Chen~Change
  Loy.
\newblock Mocanet: Motion retargeting in-the-wild via canonicalization
  networks.
\newblock In {\em AAAI}, 2022.

\end{thebibliography}

\clearpage
\appendix
\section{Appendix}
\subsection{Experimental Details}
\label{subsec:details}

\vspace{1ex}\noindent\textbf{Setup.} 
We implement the proposed model with PyTorch~\cite{NEURIPS2019_9015}. All the experiments are conducted on a Linux machine with $8$ NVIDIA V100 GPUs, which is intended for accelerating pretraining. A single GPU is usually sufficient for finetuning and inference.

\vspace{1ex}\noindent\textbf{Pretraining.} For AMASS~\cite{AMASS:2019}, we first render the parameterized human model SMPL+H~\cite{MANO:SIGGRAPHASIA:2017}, then extract 3D keypoints with a pre-defined regression matrix~\cite{bogo2016keep}. We extract 3D keypoints from Human3.6M~\cite{h36m_pami} by camera projection~\cite{lcn-pami}. We sample motion clips with length $T=243$ for 3D mocap data. For 2D data, we utilize the provided annotations of PoseTrack~\cite{PoseTrack}. We further include 2D motion from an unannotated video dataset InstaVariety~\cite{humanMotionKanazawa19} extracted by OpenPose~\cite{cao2018openpose}. Since the valid sequence lengths for in-the-wild videos are much shorter, we use $T=30$ (PoseTrack) and $T=81$ (InstaVariety). 
We convert keypoints of 2D datasets (COCO and OpenPose format) to Human3.6M using permutation and interpolation following previous works.
We set the input channels $C_\text{in} = 3$ ($x$, $y$ coordinates and confidence) following ~\cite{yan2018spatial, duanrevisiting}. Random horizontal flipping is applied as data augmentation. The whole network is trained for $90$ epochs with learning rate $0.0005$ and batch size $64$ using an Adam~\cite{kingma2014adam} optimizer. The weights of the loss terms are $\lambda_\text{O}=20$. We set the 2D skeleton masking ratio $=15\%$, same as BERT~\cite{devlin-etal-2019-bert}. More specifically, we use $10\%$ frame-level masks and $5\%$ joint-level masks. We vary the proportion of the two types of masks and only observe marginal differences. We follow ~\cite{chang2019poselifter} to fit a mixture of distributions and sample per-joint noises from it. To keep the noise smooth and avoid severe jittering, we first sample the noise $\bm{z}\in\mathbb{R}^{T_{K} \times J}$ for $T_{K}=27$ keyframes, then upsample it to $\bm{z}'\in\mathbb{R}^{T \times J}$, and add a small gaussian noise $\mathcal{N}(0, 0.002^2)$. 

\vspace{1ex}\noindent\textbf{3D Pose Estimation.} The 2D skeletons are provided by 2D pose estimator trained on MPII~\cite{andriluka14benchmark} and Human3.6M~\cite{h36m_pami} following the common practice~\cite{lcn-pami,pavllo20193d}. For training from scratch, we train for $60$ epochs with learning rate $0.0002$ and batch size $32$. For finetuning, we load the pretrained weights and train for $30$ epochs.

\vspace{1ex}\noindent\textbf{Skeleton-based Action Recognition.} We use the 2D skeleton sequences extracted with HRNet~\cite{sun2019deep} provided by PYSKL~\cite{duan2022PYSKL}. We then upsample the skeleton sequences to a uniform length $T=243$. For NTU-RGB+D, the DSTformer output after global average pooling is fed into an MLP including dropout $p=0.5$, a hidden layer of $2048$ channels, BatchNorm, and ReLU. We train for $200$ epochs with learning rate $0.001$ and batch size $32$ for training from scratch. We set learning rate $0.0001$ for the backbone, and learning rate $0.001$ for the downstream MLP for finetuning. For one-shot recognition on NTU-RGB+D-120, we apply dropout $p=0.1$, and use a linear layer to get action representation of size $2048$. We train with batch size $16$, and each batch includes $8$ action pairs. Samples from the same action class are set as positives against the negatives from the remainder of the batch. We use the supervised contrastive loss~\cite{SupContrast} with temperature $0.1$.

\vspace{1ex}\noindent\textbf{Human Mesh Recovery.}
For the Human3.6M dataset, we use the same 2D skeleton sequences as the 3D pose estimation task. The SMPL ground-truth (GT) parameters are obtained by MoSh~\cite{loper2014mosh} which are fitted to the physical mocap markers. For the 3DPW dataset, we obtain its detected 2D skeleton sequences from DarkPose~\cite{Zhang_2020_CVPR} provided by~\cite{choi2020pose2mesh}. The SMPL GT parameters of 3DPW are obtained using IMUs~\cite{vonMarcard2018}. Following previous work \cite{kanazawa2018end, kocabas2020vibe, wan2021}, we add COCO~\cite{lin2014microsoft} datasets for training, following~\cite{choi2020pose2mesh}. The pseudo-GT SMPL parameters of COCO are provided by EFT~\cite{joo2021exemplar}. 
Following~\cite{kolotouros2019learning, kocabas2020vibe}, we use the 6D continuous rotation representations~\cite{zhou2019continuity} instead of original axis angles when estimating pose parameters. The shape and pose MLPs are the same, including one hidden layer of $512$ channels, BatchNorm, and ReLU. We sample motion clips with length $T=16$ and stride $8$ for both datasets. The weights of the loss terms are $\lambda^{\text{m}}_\text{3D}=0.5$, $\lambda_{\theta}=1000$, $\lambda_{\beta}=1$, $\lambda_{\text{n}}=20$ and $\lambda^{\text{m}}_\text{O}=10$. We train the whole network end-to-end for $60$ epochs with batch size $128$, with learning rates $0.00005$ for the backbone, and $0.0005$ for the downstream network, respectively.  %

\subsection{Additional Experiments}

We first provide a deeper analysis of the proposed model architecture, then present more results on skeleton-based action recognition. We also include an ablation study on mask ratio for pertaining as well as the qualitative results.

\subsubsection{Analysis of DSTformer}

We investigate the proposed DSTformer model to interpret its performance improvement and re-evaluate the design assumptions in Section 3.2. The major difference between DSTformer and existing works~\cite{zheng20213d, li2022mhformer, zhang2022mixste} lies in the dual-stream design and the adaptive fusion mechanism. Therefore, we probe the fusion weights $\bm{\alpha}_{\text{ST}}$ and $\bm{\alpha}_{\text{TS}}$ produced by the attention regressor to understand how DSTformer balances the two streams and adapts to different inputs. 

\begin{figure}[t]
 
  \centering
  \includegraphics[width=\linewidth]{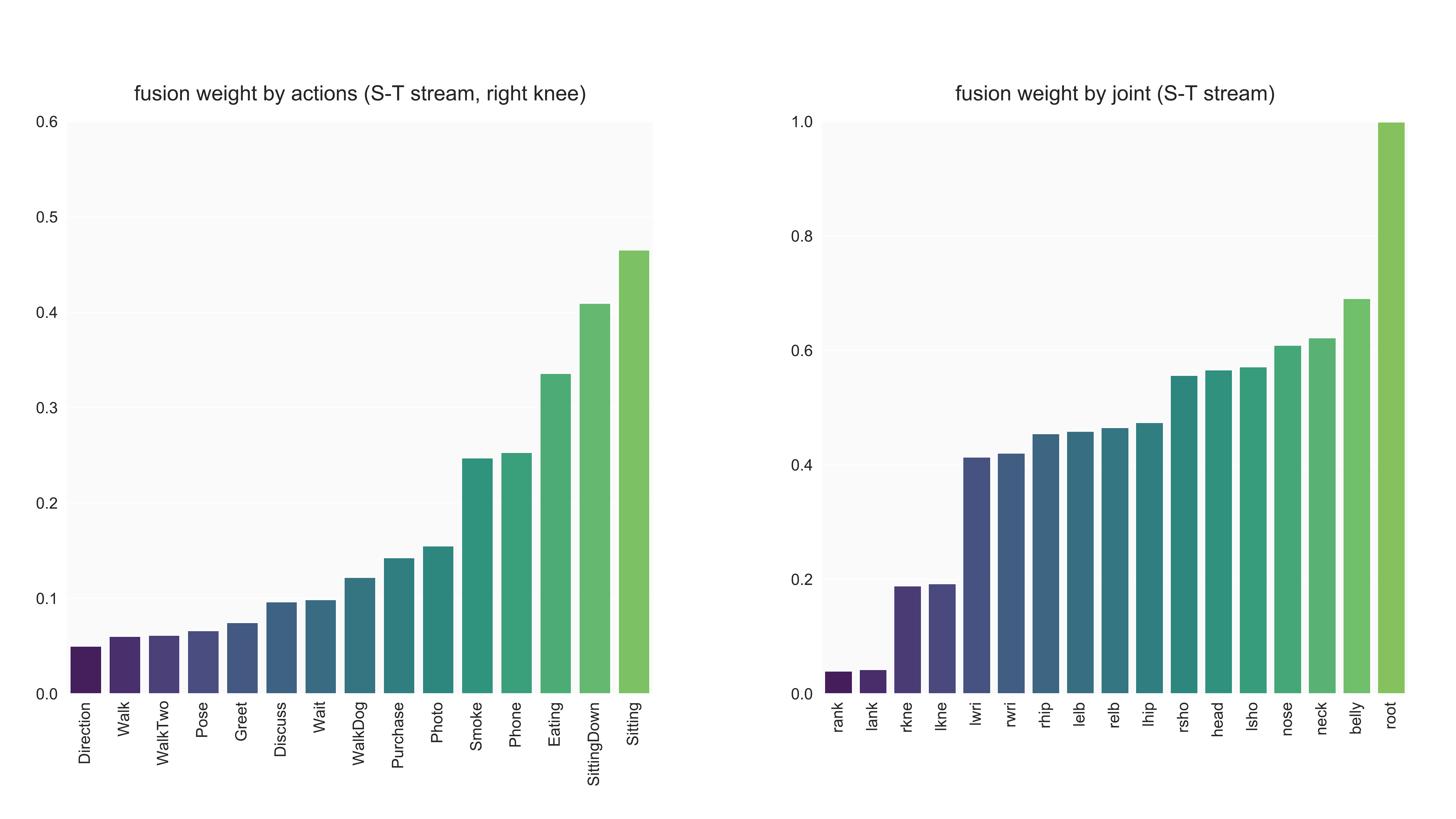}
  \caption{\textbf{Statistic of fusion weights grouped by actions (left) and joints (right).} We plot the average fusion weights for the S-T stream on the Human3.6M test set. The third layer of a 3D pose estimation model trained from scratch on Human3.6M is used for probing.}
  \label{fig:dstformer_actions}
\end{figure}

We first examine how different input actions influence the dual-stream fusion weights, and the results are shown in Figure~\ref{fig:dstformer_actions} (left). Note that $\bm{\alpha}_{\text{ST}} + \bm{\alpha}_{\text{TS}} = \mathbf{1}$, so we only analyze $\bm{\alpha}_{\text{ST}}$ for simplicity. We find that actions with a higher degree of movement (\eg walking) lead to a smaller fusion weight for the \emph{S-T} stream, which means the model is inclined to the \emph{T-S} stream. Oppositely, when the input is from a more static action (\eg sitting and eating), the model tends to increase the weights of the \emph{S-T} stream. These results align with our assumption that the two streams could focus on different spatial-temporal aspects of features and that the model can reweigh the two streams based on the input characteristics. Intuitively, the \emph{S-T} branch is more specialized for spatial modeling, while the \emph{T-S} branch is more focused on temporal modeling. 

\begin{figure}[t]
 
  \centering
  \includegraphics[width=\linewidth]{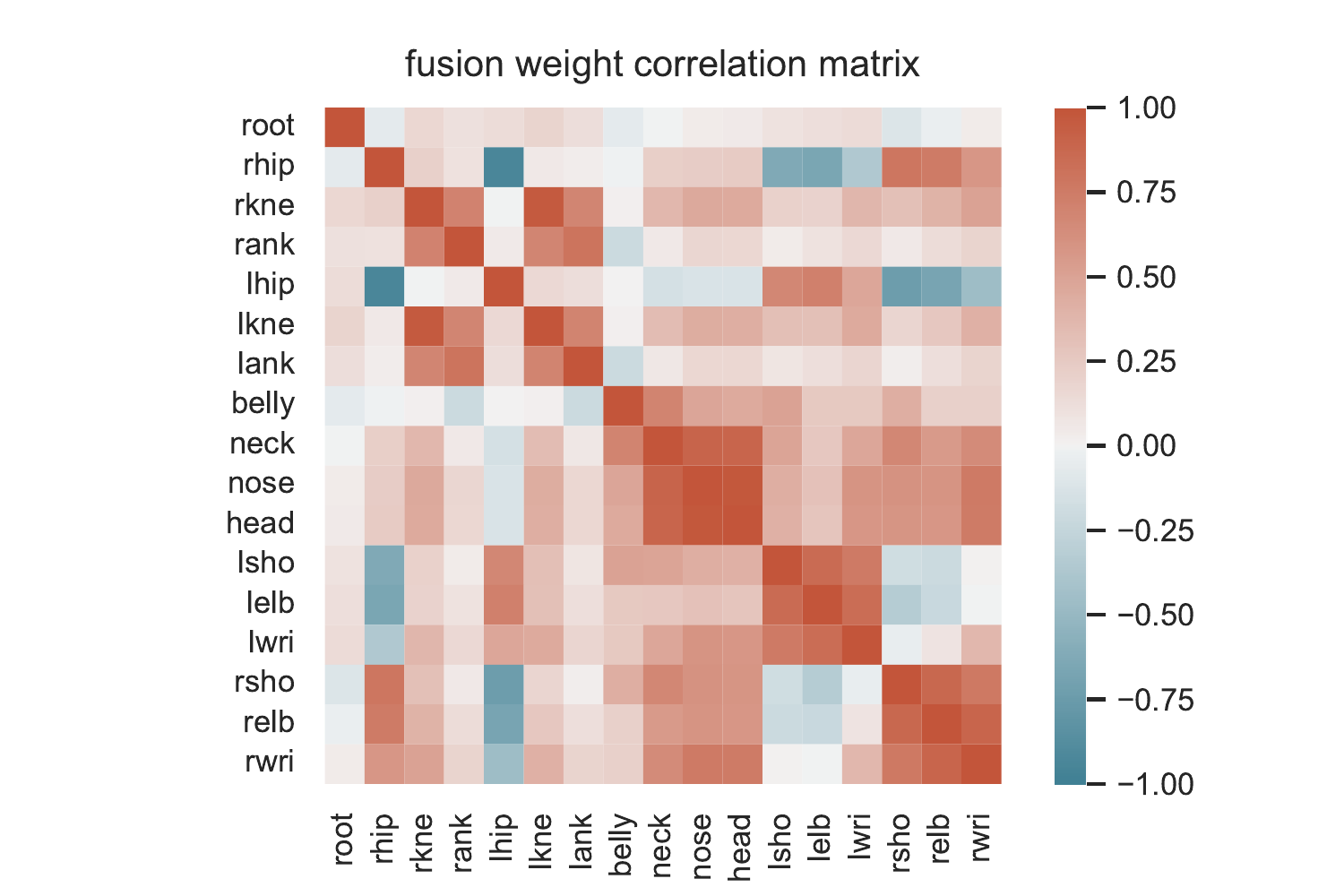}
  \caption{\textbf{Correlation matrix of the joint-wise fusion weights.} The results are computed on the Human3.6M test set. We use the third layer of a 3D pose estimation model trained from scratch on Human3.6M for probing.}
  \label{fig:dstformer_correlation}
\end{figure}

We further study the distribution of fusion weights across the body joints. As illustrated in Figure~\ref{fig:dstformer_actions} (right), the fusion weights vary dramatically across different body joints. Specifically, the model leans towards the \emph{T-S} branch for keypoints with large temporal variations (limb joints), while it gives more importance to the \emph{S-T} branch for keypoints that are more stable (torso joints). The result also corresponds to our ``division and cooperation'' design of the two streams. In addition, we analyze the correlation of fusion weights for different body joints. Figure~\ref{fig:dstformer_correlation} indicates that fusion weights of joints from the same functional part (\eg arms, legs, head, torso) are naturally correlated, which implies that the model learns the joint connections implicitly and adjusts their weights in an interrelated manner with regard to input changes.

\begin{table}[t]
\begin{center}
\setlength{\tabcolsep}{3pt}
\resizebox{0.7\linewidth}{!}{
\small
\begin{tabular}{ccc|c}
\thickhline 
Depth($N$) & Heads($h$) & Channels($C_{\text{e}}$)& MPJPE  \\

\thickhline
4 & 8 & 512 & 39.7  \\
6 & 8 & 512 & 40.4  \\
5 & 6 & 512 & 39.6 \\
5 & 10 & 512 & 39.5  \\
5 & 8 & 256 & 40.4  \\
5 & 8 & 512 & \textbf{39.2}  \\
\thickhline 
\end{tabular}
}
\end{center}
\caption{\textbf{Comparison of model architecture design.} All the methods are trained on Human3.6M from scratch and measured by MPJPE (mm).}
\label{tab:ablation_arch}
\end{table}

\subsubsection{Architecture Hyperparameter Analysis}
We explore how different architecture hyperparameters influence the performance of DSTformer. In Table \ref{tab:ablation_arch}, we vary the number of stacking modules (\emph{depth}), self-attention heads (\emph{heads}), and embedding size (\emph{channels}) and measure their performance on the 3D human pose estimation task. We then select the optimal combination and apply the design to all the tasks without additional adjustment.

\begin{table}[h]

\center
\small
\resizebox{\linewidth}{!}{
\begin{tabular}{l  | c  c  c | c  c  c }
\thickhline 
\multirow{2}{*}{Method} & \multicolumn{3}{c|}{X-Sub} & \multicolumn{3}{c}{X-View}  \\
\cline{2-7}
& scratch  & finetune  & diff  & scratch & finetune  & diff  \\
\hline 

MS$^{2}$L~\cite{lin2020ms2l} & 78.4 & 78.6 & +0.2 & - & - & - \\
CrosSCLR~\cite{li2021crossclr} & 85.2 & 86.2 & +1.0 & 91.4 & 92.5 & +1.1  \\
Skel-Con~\cite{Thoker2021skeleton} & 72.9 & 79.3 & \textbf{+6.4} & 87.8 & 90.4 & +2.6  \\
MCC~\cite{Su_2021_ICCV} & \textbf{88.5} & 89.7 & +1.2 & \textbf{95.1} & 96.3 & +1.2  \\
SCC~\cite{Yang_2021_ICCV} & - & 88.0 & - & - & 94.9 & -  \\
\hline
\rowcolor{mygray}
Ours & 87.7 & \textbf{93.0} & +5.3 & 94.1 & \textbf{97.2} & \textbf{+3.1}  \\

\thickhline

\end{tabular}}
\vspace{0.2cm}
\caption{\textbf{Quantitative comparison with self-supervised action recognition approaches.} We measure the top-1 accuracy on NTU-RGB+D. \emph{diff} measures the performance gap between \emph{scratch} and \emph{finetune}.
}
\label{tab:action_ssl}
\end{table}

\subsubsection{Comparison with Self-supervised Action Recognition Approaches}
Our framework also shares some commonalities with the research direction of self-supervised skeleton-based action recognition ~\cite{lin2020ms2l, Su_2021_ICCV, Thoker2021skeleton, li2021crossclr, Yang_2021_ICCV}. These methods design various pretext tasks on 3D skeleton sequences and learn 3D motion representations without action labels (\ie self-supervised learning). To our best knowledge, our method is the first to use 2D-to-3D lifting as the pretext task for skeleton-based action recognition. In this way, it learns 3D-aware motion representations on 2D skeleton sequences. While previous methods learn 3D motion representations that require 3D skeleton sequences as input (usually obtained with RGB-D devices such as Azure Kinect), the 2D motion representations can be applied to in-the-wild videos (from mobile devices or the Internet) more easily. We conduct a quantitative comparison and show the results in Table~\ref{tab:action_ssl}. The accuracy gain of finetuning compared to training from scratch could be considered as the contribution of learned motion representations. Although our network is not specifically designed for the action recognition task (\emph{scratch}), it surpasses the previous methods on fine-tuned performances (\emph{finetune}) and shows considerable performance gains (\emph{diff}).

\begin{figure}[t]
 
  \centering
  \includegraphics[width=\linewidth]{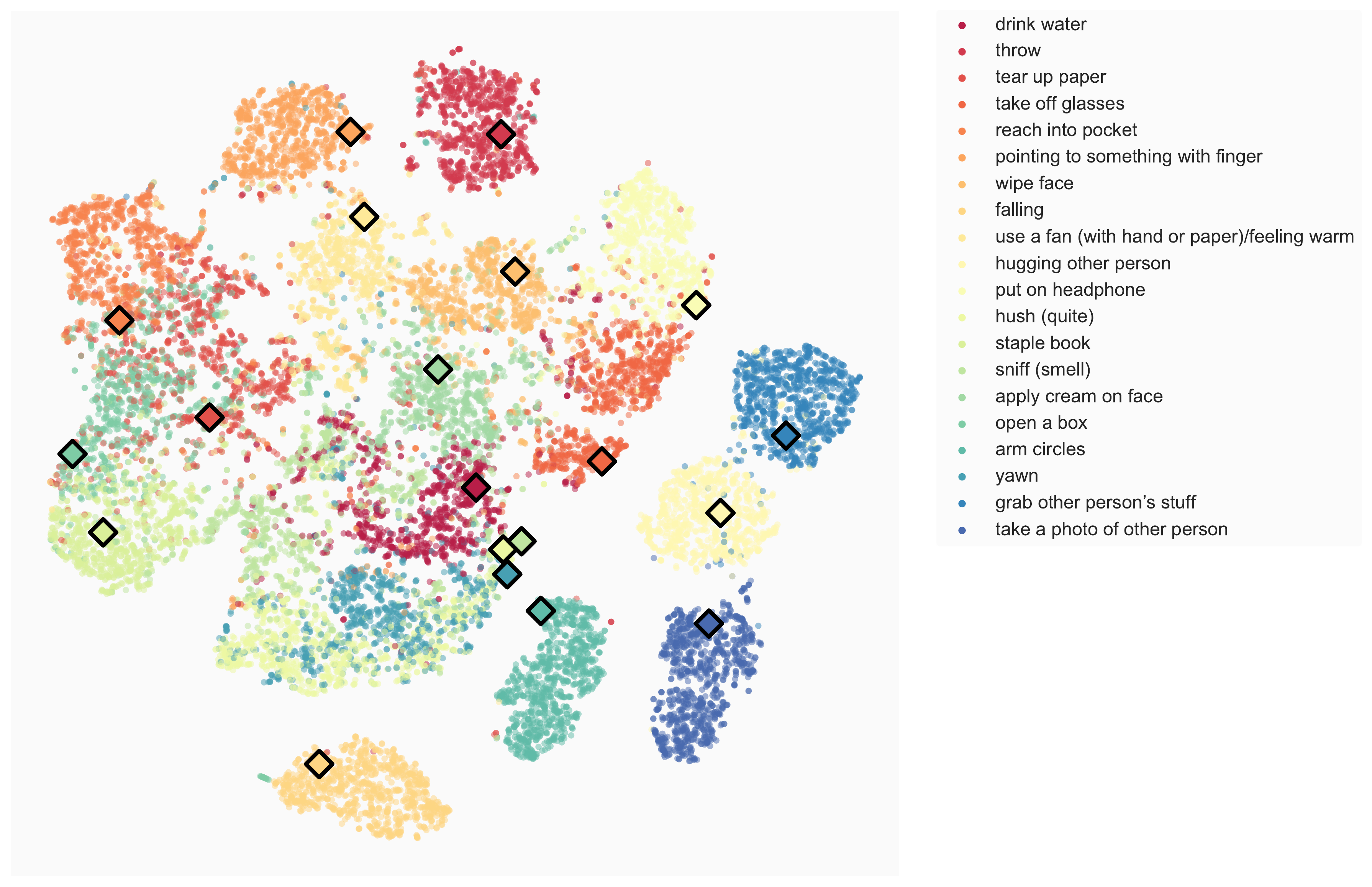}
  \caption{\textbf{Visualization of action representations for one-shot recognition.} We show the action representations of the $20$ novel classes after training on the other $100$ auxiliary classes with contrastive learning. We compute the pairwise cosine distance and apply t-SNE. Diamonds indicate the labeled exemplars for one-shot recognition.}
  \label{fig:oneshot}
\end{figure}

\subsubsection{One-shot Action Recognition}

Figure~\ref{fig:oneshot} shows the learned action representation for the novel classes with our finetuned model. We can see that meaningful clusters emerge although the model has never seen these actions before. We perform one-shot action recognition using 1-nearest-neighbor with the exemplars and achieve $67.4\%$ top-1 accuracy.

\subsubsection{Mask Ratio Analysis}

\begin{figure}[t]
  \centering
  \includegraphics[width=0.8\linewidth]{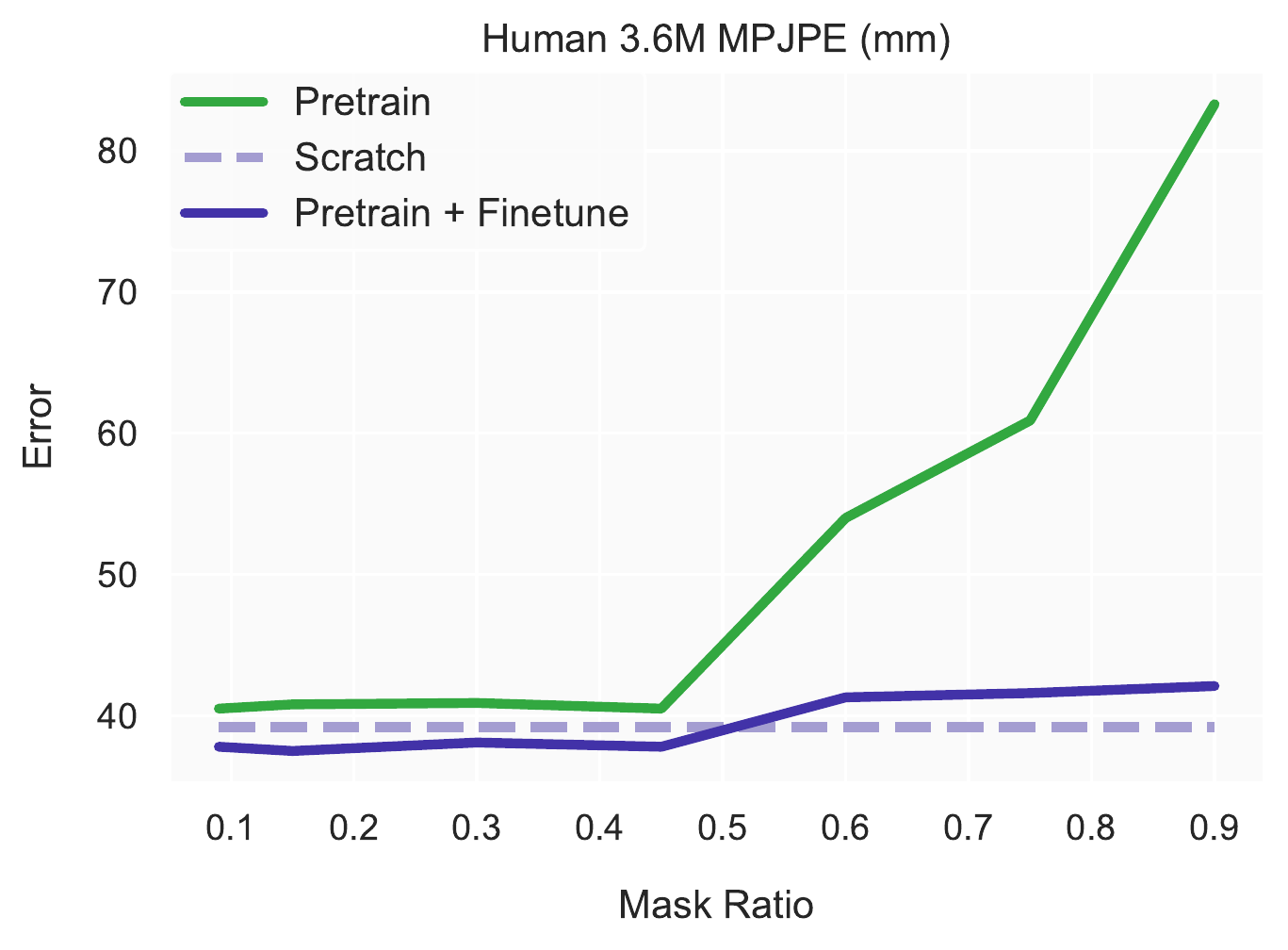}
\caption{\textbf{Ablation study of pretraining mask ratios.} For each mask ratio, We test the Human3.6M MPJPE of the pretrained models with/without finetuning respectively.}
  \label{fig:ablation_mask}
\end{figure}

We study the effect of different masking ratios on both pretraining and downstream finetuning. We measure the performance by MPJPE on the Human3.6M test set. Frame-level mask and joint-level mask are fixed to $2:1$. As Figure~\ref{fig:ablation_mask} shows, when the masking ratio is less than $45\%$, the pretrained models could learn 2D-to-3D lifting well, and pretraining improves the finetuning performance (better than the scratch baseline). As we continue to increase the masking ratio, the pretrained models perform much worse at 2D-to-3D lifting, and pretraining is no longer beneficial for the finetuning performance. Our observation is slightly different from MAE~\cite{he2021mae} where a high masking ratio ($75\%$) works well. One possible explanation is that skeletons are highly abstract representations of human motion. A high masking ratio leaves too few cues to complete the whole sequence, therefore the model tends to overfit. It is also worth noting that in our pretraining task (2D-to-3D lifting), the depth channel ($\frac{1}{3}$ of original data) is intrinsically masked.

\subsection{Limitations}
The primary limitation of this work lies in its focus on single-person skeleton sequences, which results in the motion representation being insensitive to factors such as appearance, surroundings, and interactions.
To overcome these limitations, future work could explore combining the learned motion representation with generic video features or scene representations. 

\end{document}